\documentclass[journal]{IEEEtran}

\usepackage{cite}
\usepackage{amsmath,amssymb,amsfonts}
\usepackage{bm}
\usepackage{graphicx}
\usepackage{booktabs}
\usepackage{multirow}
\usepackage{array}
\usepackage{tabularx}
\usepackage{threeparttable}
\usepackage{url}
\usepackage{balance}
\usepackage{hyperref}
\hypersetup{
    colorlinks=true,
    linkcolor=blue,
    citecolor=blue,
    urlcolor=blue
}

\usepackage{xcolor}
\definecolor{RevisionBlue}{RGB}{0,70,180}
\newif\ifmarkchanges
\markchangesfalse
\DeclareRobustCommand{\rev}[1]{\ifmarkchanges{\color{RevisionBlue}#1}\else#1\fi}
\AtBeginDocument{%
  \setlength{\abovedisplayskip}{3pt plus 1pt minus 1pt}%
  \setlength{\belowdisplayskip}{3pt plus 1pt minus 1pt}%
  \setlength{\abovedisplayshortskip}{2pt plus 1pt minus 1pt}%
  \setlength{\belowdisplayshortskip}{2pt plus 1pt minus 1pt}%
  \setlength{\jot}{2pt}%
  \raggedbottom
}

\begin{document}

\title{SPOC-Net: Single-Primitive Online Composition Network for GNSS Jamming Set Recognition}

\author{Zhihan Zeng, \IEEEmembership{Graduate Student Member, IEEE}, Kaihe Wang, \IEEEmembership{Graduate Student Member, IEEE},\\ José A. López-Salcedo, \IEEEmembership{Senior Member, IEEE}, Gonzalo Seco-Granados, \IEEEmembership{Fellow, IEEE}, Zhongpei Zhang 

\thanks{Zhihan Zeng is with the National Key Laboratory of Wireless Communications, University of Electronic Science and Technology of China (UESTC), Chengdu, China (e-mail: 202511220608@std.uestc.edu.cn). Kaihe Wang and Zhongpei Zhang are with the Shenzhen Institute for Advanced Study, UESTC, Shenzhen 518110, China (e-mail: khewang@yeah.net; zhangzp@uestc.edu.cn). José A. López-Salcedo and Gonzalo Seco-Granados are with the Universitat Autònoma de
Barcelona, 08193 Barcelona, Spain, E-mail: (jose.salcedo@uab.cat, gonzalo.seco@uab.cat). The corresponding author is Zhongpei Zhang.
}
}
\bstctlcite{IEEEexample:BSTcontrol}
\maketitle

\begin{abstract}
\rev{Reliable positioning, navigation, and timing support intelligent transportation, autonomous systems, and space-air-ground integrated networks. However, global navigation satellite system (GNSS) jamming recognizers that treat each mixture as a separate class are difficult to extend to new combinations. Therefore, this paper proposes SPOC-Net, which decomposes the recognition problem into identifying a set of basic jamming components. Multi-resolution time-frequency features and learned component queries provide evidence for each component type. A high-resolution branch estimates the number of active types, and a structured decoder combines this estimate with component evidence to select a valid set. For training, measured single-component records are the only physical samples used in gradient optimization. Their associated clean in-phase and quadrature (IQ) sequences are combined on demand during training to produce labeled mixtures with different relative powers and jamming-to-noise ratios. Separate measured mixtures from ten training-listed compositions support model selection and decoder calibration; six other compositions are reserved for final testing. Evaluation on 14,220 independently generated, conductively combined, and recorded radio frequency mixtures yields 80.69\% exact-set accuracy and a 92.84\% micro-averaged F1 score. On combinations excluded from model development, SPOC-Net achieves 80.89\% exact-set accuracy, exceeding the strongest comparison method by 18.77 percentage points under the reported protocols.}

\end{abstract}


\section{Introduction}
\IEEEPARstart{T}{he} International Mobile Telecommunications 2030 framework is expected to support ubiquitous connectivity, integrated sensing and communication, high-accuracy positioning, and stronger security and resilience \cite{ITU2023IMT2030}. Space-air-ground integrated networks (SAGINs) connect satellites, aerial platforms, and terrestrial systems to extend services over land, sea, air, and remote regions \cite{Cui2022SAGIN}. These networks depend on reliable positioning, navigation, and timing for synchronization, mobility control, autonomous operation, and emergency services. \rev{To support these functions, many civil and industrial systems rely on the global navigation satellite system (GNSS).}

A GNSS receiver observes satellite signals after severe propagation loss, and the useful signal is normally close to or below the receiver noise floor before correlation. A nearby jammer can therefore dominate the receiver front end and interrupt acquisition, tracking, or navigation with limited transmitted power \cite{Gao2016JPROC_Protecting,Wesson2018TAES_PowerDistortion}. Reliable jamming recognition is needed before a receiver or monitoring system can select an appropriate suppression or monitoring method.

\rev{In this context, a receiver must contend with several possible jamming signals, including single-tone, multitone, chirp, pulsed, and partial-band noise interference. Each individual jamming type is termed a \emph{primitive}, and a \emph{composition} is a set of primitive types that are simultaneously active. With $C$ primitive types, there are up to $2^{C}-1$ nonempty compositions. Even when mixed-jamming recognition is restricted to two or three active types, the number of combinations requiring labeled training data grows rapidly. In a controlled dataset-collection campaign, acquiring each mixture requires coordinated signal generation, controlled component powers, repeated receiver captures, and verified labels. These requirements concern the construction of a reproducible labeled dataset; they do not imply that real-world jammers are synchronized or have controlled power relations. The resulting measurement effort makes comprehensive mixture collection difficult.}

\rev{Most learning-based GNSS jamming methods use a closed class list and assign one softmax label to each single or mixed type \cite{Chen2022TAES_FingerprintDNN,Zhong2024TIM_TSFANet,Mehr2025TAES_DNNGNSS,Jia2025TAES_MSFFKAN,Jiang2026TCCN_ACSNet,Xiao2025TAES_CompoundFusion}. Such methods can perform well for a fixed taxonomy, but accommodating an additional combination generally requires extending the class list and retraining the classifier. A component-wise output instead identifies the active primitive types and can represent different combinations without defining a separate output neuron for each mixture. Multi-label compound recognition and generalization to combinations excluded from training have already been studied in radar jamming \cite{Meng2023RS_MultiLabelCompound,Xiao2024IET_OpenCompound, zeng2026gackan, zeng2026phygmoe, zeng2026skanet, zeng2026jsrgfnet}.}

\rev{This paper proposes the Single-Primitive Online Composition Network for GNSS Jamming Set Recognition (SPOC-Net), a recognition method that learns to identify mixed jamming while using only measured single-component records for gradient optimization. A record containing one primitive type is called a \emph{singleton}. The associated clean in-phase and quadrature (IQ) sequences are combined before the short-time Fourier transform (STFT) to create additional labeled training examples. Here, \emph{online composition} means generating a fresh mixture from stored singleton IQ sequences when the training loader requests it, rather than constructing a fixed mixed dataset in advance; it does not mean real-time radio frequency (RF) acquisition or signal combination during inference. The resulting \emph{auxiliary} examples supplement measured singletons, and their known component labels provide the training targets, or \emph{supervision}. Relative component powers and aggregate jamming-to-noise ratio (JNR) are sampled by this training-data generator.}

\rev{SPOC-Net receives an STFT image, not a JNR value or acquisition metadata: the sampled JNR controls auxiliary-signal generation, whereas the recorded JNR labels organize the experimental results. Ten compositions are permitted in auxiliary training and are termed \emph{training-listed}. Separate measured mixtures of these same ten compositions are used for model selection and decoder calibration. Six other combinations of the known primitives are \emph{held out}: their complete component sets are excluded from auxiliary training and all model-development decisions, and their measured records are used only for final testing. This distinction tests generalization to new combinations of known types, not recognition of an unknown jamming type. The evaluated task contains two-component and three-component mixtures; singleton records additionally support an auxiliary training task that distinguishes one component from a mixture.}

The contributions of this paper are summarized below.
\begin{enumerate}
    \item \rev{An online clean-IQ composer supplies labeled mixed-jamming examples without requiring measured mixtures for gradient updates. It varies component powers, aggregate JNR, and background noise before the STFT, and forms related two-component and three-component examples to train component-count estimation.}
    \item \rev{A new recognition method, referred to as SPOC-Net, combines a coordinate-aware multi-resolution encoder, learned primitive queries, a detached high-resolution cardinality branch, and structured candidate-set scoring. Cardinality denotes the number of active primitive types. The model shares component evidence across compositions and jointly selects component identities and a valid set size.}
    \item \rev{A conducted RF evaluation platform and an explicitly separated development--test protocol assess transfer from singleton-based training to measured mixtures. The platform combines a live GNSS signal with up to three generated jammer signals. Entire held-out compositions, not merely individual recordings, are excluded from gradient training, checkpoint selection, hyperparameter tuning, and decoder calibration.}
\end{enumerate}

\rev{On 14,220 measured mixed records, SPOC-Net achieves 80.69\% exact-set accuracy and 92.84\% micro-averaged F1 (micro-F1). Its held-out exact-set accuracy is 80.89\%, compared with 62.12\% for the strongest comparison method. These results assess the complete training and inference procedures described below; they are not an architecture-only comparison.}

\rev{The remainder of this paper is organized as follows. Section~II reviews related work. Section~\ref{sec:system} defines the received-signal model, primitive waveforms, recognition task, and data-use protocol. Section~\ref{sec:method} presents the complete SPOC-Net method, from signal representation and online composition to network architecture, learning, and decoding. Section~\ref{sec:experiments} describes the experimental setup and comparison protocols. Section~\ref{sec:results} analyzes the results, and Section~\ref{sec:conclusion} concludes the paper.}

\section{Related Work}
\label{sec:related}

\subsection{Conventional GNSS Interference Detection and Mitigation}

GNSS interference protection usually includes detection and mitigation. Before correlation, useful satellite signals are close to or below the noise floor, so interference detection often relies on changes in received power, signal statistics, or patterns in the time-frequency plane. Statistical methods can detect both stationary and nonstationary interference \cite{Wang2017TAES_TFStat,Wang2018TAES_TF_GNSS}, while power and distortion monitoring can provide receiver-level evidence of abnormal signals \cite{Wesson2018TAES_PowerDistortion}. These approaches are interpretable and usually do not require a large labeled dataset, but they mainly determine whether interference exists and do not always identify every active component in a compound record.

Mitigation methods are commonly matched to signal structure. Adaptive notch filters are widely used for narrowband interference \cite{Borio2014ICLGNSS_MultiStateNotch,Gamba2019ICLGNSS_FLLNotch,Qin2020TAES_ANF, Zeng2026ICCC, Zeng2026CrossHeightCKM, Zeng2026QuaMoEDRF, Zeng2026GeoUQGFNet}, while subband gain control provides frequency-selective suppression \cite{Garzia2021ICLGNSS_SubbandAGC}. Chirp interference can be processed through fractional Fourier and Zak transform methods \cite{Qin2022TAES_ChirpSA,Sun2024TAES_FrFT,Luo2024TAES_Zak}. Nonnegative matrix factorization can separate structured components in a time-frequency representation \cite{Silva2022TAES_NMFDetection,Silva2023TAES_NMFMitigation}, and antenna-array or polarization methods can exploit spatial and polarization information \cite{Li2025TAES_Multipolarization}. Since each method depends on a particular signal feature, reliable component recognition is needed before an appropriate suppression method can be selected.

\subsection{Learning-Based GNSS Interference Recognition}

Machine learning has been widely studied for GNSS interference recognition. Spectra, spectrograms, and spectrum-waterfall images are common inputs because they expose both frequency positions and temporal variations. Support vector machines and convolutional neural networks have been used to classify interference types from these representations \cite{Ferre2019Sensors_Jammer,Cai2019ICCC_Waterfall,Mehr2022ICLGNSS_CNN}. Fingerprint-spectrum networks and temporal-spatial feature aggregation further improve the use of spectral and local pattern information \cite{Chen2022TAES_FingerprintDNN,Zhong2024TIM_TSFANet}.

Recent studies also address limited training data, low computational cost, and difficult JNR conditions. Low-resource classifiers and lightweight networks have been developed for practical GNSS receivers \cite{VanderMerwe2024TAES_LowResource,Mehr2025TAES_DNNGNSS,Jia2025TAES_MSFFKAN,Jia2025TIM_LowPowerCNN}. Graph-convolution models provide another way to describe relations among signal features \cite{Li2025JSEN_DualGCN}. Interpretable models, recurrent networks, and few-shot learning have also been studied to improve decision transparency, recognition speed, and adaptation ability \cite{Jeeru2025JSEN_Interpretable,Mehr2025PLANS_GRU,Ott2024ICLGNSS_FewShot}. Most existing methods still use closed-set classification, in which each observation receives one class or every predefined compound type is treated as an independent class. This setting is effective when training covers all test classes, \rev{but it does not directly identify the active jamming types.} As the primitive vocabulary and component count increase, the number of compound classes and the demand for measured training records also increase.

\subsection{Compound Interference Recognition and Physical Data}

Compound GNSS interference recognition has received increasing attention. Object-detection methods have been used to locate several interference patterns in one time-frequency image \cite{Liu2024ICCASIT_YOLO}, and dedicated convolutional networks have been designed for compound classification \cite{Jiang2026TCCN_ACSNet}. Time-frequency features have also been combined with power-spectrum features to improve recognition of overlapping signals \cite{Xiao2025TAES_CompoundFusion}. High-resolution interference sensing has been coupled with neural networks for multiple-interference processing \cite{Song2026JSEN_TFUNet}. These studies confirm that mixed observations contain useful component information, although their training protocols commonly depend on a fixed compound list or measured compound records.

\rev{Physical data collection creates an additional constraint. A controlled single-jammer capture is relatively straightforward to label, whereas collecting a reproducible mixture requires multiple sources, specified power relations, repeated captures, and checks of receiver behavior. The RF hardware can introduce channel-dependent responses, combiner loss, gain changes, and nonlinear effects that numerical signal addition does not reproduce completely. Measured GNSS interference, low-resource recognition, and few-shot adaptation have been studied in recent work \cite{Mehr2022ICLGNSS_CNN,Ott2024ICLGNSS_FewShot,Spanghero2025TIFS_JammerLocalization,Jia2025TIM_LowPowerCNN}.}

\subsection{Compositional Generalization and Structured Set Prediction}

Multi-label learning assigns one binary output to each primitive and \rev{directly identifies the active jamming types in a mixed record} \cite{Zhang2014TKDE_MultiLabelReview}. It also allows one primitive representation to be reused across several compositions. Meng et al. used complex-valued multi-label learning for simulated radar compound jamming \cite{Meng2023RS_MultiLabelCompound}. Xiao et al. studied simulated radar compositions excluded from training and combined multi-label classification with image reconstruction and extreme-value modeling for unknown-primitive detection \cite{Xiao2024IET_OpenCompound}. These studies establish the value of component-wise outputs for compound signals.

\rev{Query-based decoders assign a learned query to each label and collect label-related evidence from a shared feature map. Attention provides the basic operation \cite{Vaswani2017NIPS_Attention}, while Query2Label and ML-Decoder show how label queries support multi-label (ML) classification \cite{Liu2021Q2L,Ridnik2023WACV_MLDecoder}. The asymmetric loss (ASL) reduces the influence of easy negative labels \cite{Ridnik2021ICCV_ASL}. Independent binary decisions, however, do not constrain the number of selected components or enforce an admissible combination.}

\rev{The present work brings these ideas together for GNSS jamming recognition when measured mixtures are excluded from gradient optimization. It combines online complex-IQ composition with separate primitive-evidence and cardinality branches and a decoder that scores complete valid sets. Independently acquired conducted RF mixtures assess whether the resulting recognizer transfers to combinations omitted from training and model development. This setting concerns new combinations of known primitives; unknown-primitive rejection is outside its scope.}

\section{System Model}
\label{sec:system}

\subsection{Received-Signal Formulation}

\begin{figure}[t]
\centering
\includegraphics[width=\columnwidth]{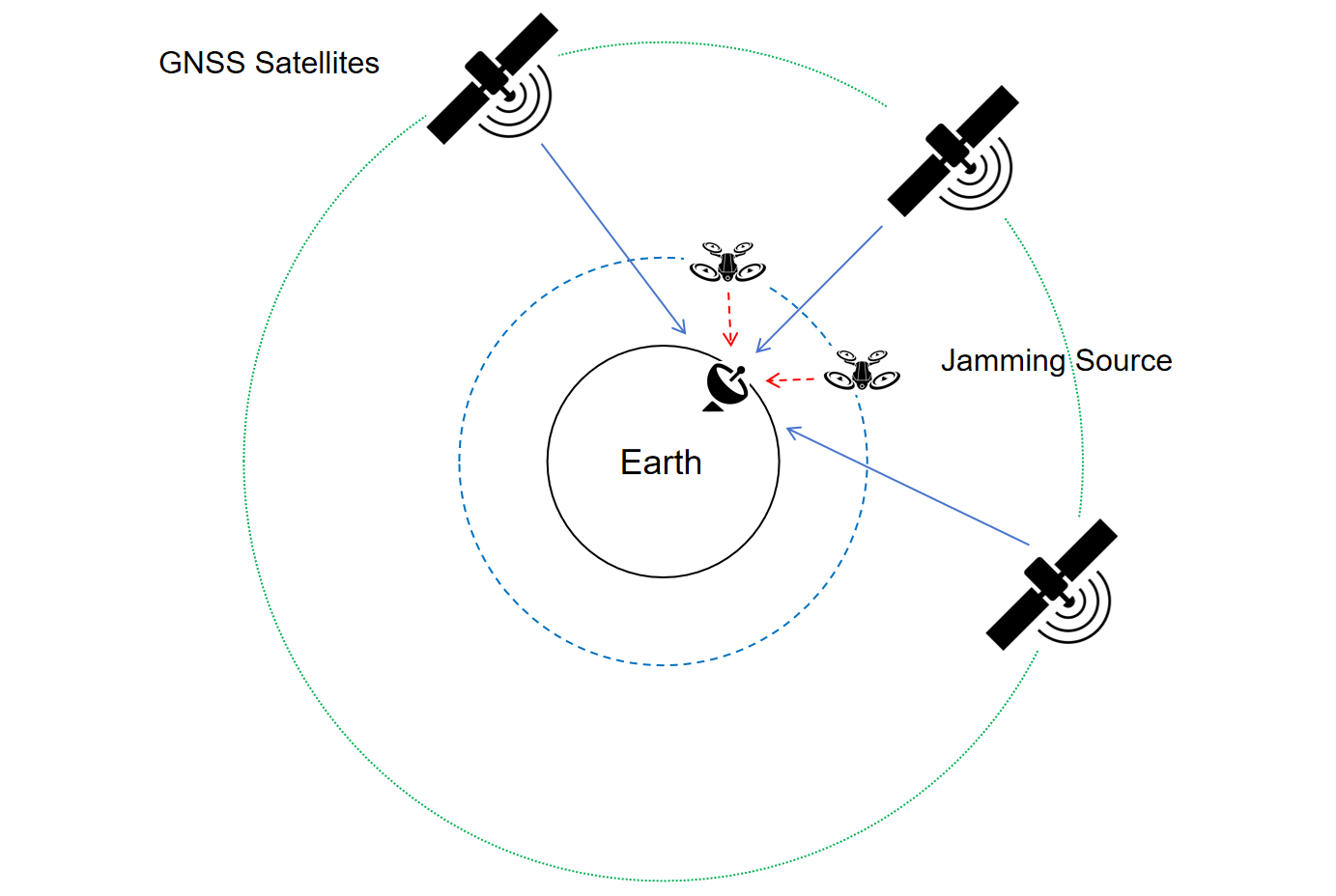}
\caption{Conceptual GNSS jamming scenario motivating the received-signal model.}
\label{fig:systemmodel}
\end{figure}

\rev{Fig.~\ref{fig:systemmodel} depicts a conceptual reception scenario in which GNSS and jamming signals arrive at a receiver. Its analog front end observes their superposition together with noise. After downconversion and sampling at frequency $F_s$, one signal snapshot, termed a \emph{record}, contains $N$ complex baseband samples. The model applies to reception through an antenna as well as conducted injection; the RF combiner used for the experiments is a particular implementation described in Section~\ref{sec:experiments}. The sampled signal and its aggregate JNR are}
\begin{subequations}
\label{eq:received_signal_and_jnr}
\begin{align}
 x[n]&=s_{\mathrm{GNSS}}[n]+\sum_{k\in\mathcal{K}}\sqrt{P_k}\,j_k[n]+w[n],
 \qquad 0\le n<N, \\
 \mathrm{JNR}&=10\log_{10}\frac{\frac{1}{N}\sum_{n=0}^{N-1}\left|\sum_{k\in\mathcal{K}}\sqrt{P_k}j_k[n]\right|^2}{\sigma_w^2}.
\end{align}
\end{subequations}
\rev{Here, $s_{\mathrm{GNSS}}[n]$ is the received GNSS signal, $\mathcal{K}$ is the active set of jamming types, $j_k[n]$ is the unit-power waveform of type $k$, and $P_k$ is its received component power. The term $w[n]$ represents thermal and receiver noise with average power $\sigma_w^2$. These definitions do not depend on whether the signals propagate over the air or pass through a conducted path. At the precorrelation stage, the spread-spectrum GNSS signal is typically below the noise floor, so the classifier primarily observes jamming time-frequency patterns in a noise-like background \cite{Gao2016JPROC_Protecting,Wang2018TAES_TF_GNSS}. For the reported implementation, $F_s=20$ MHz and each record spans $1$ ms, giving $N=20{,}000$. The evaluation covers aggregate JNR values from $-20$ to $15$ dB in 5-dB increments.}

\subsection{Primitive Waveforms}

\rev{The primitive vocabulary contains five jamming waveform types. Each realized waveform is normalized to unit average power over the record before relative component weighting.} Single-tone jamming (STJ) forms one persistent narrowband ridge and is a canonical continuous-wave threat \cite{Borio2014ICLGNSS_MultiStateNotch,Gamba2019ICLGNSS_FLLNotch}. Multitone jamming (MTJ) occupies several narrow frequency locations and can defeat a single-notch response \cite{Ferre2019Sensors_Jammer,Cai2019ICCC_Waterfall}. Linear frequency-modulated jamming (LFMJ) produces a chirp trajectory that motivates specialized time-frequency and fractional-domain processing \cite{Qin2022TAES_ChirpSA,Sun2024TAES_FrFT}. Pulsed-tone jamming (PTJ) gates a narrowband carrier and creates intermittent structures, while partial-band noise jamming (PBNJ) injects shaped noise into only part of the receiver bandwidth \cite{Garzia2021ICLGNSS_SubbandAGC,Silva2023TAES_NMFMitigation}. \rev{Their complex envelopes, before this per-record normalization, are}
\begin{subequations}
\label{eq:primitive_waveforms}
\begin{align}
 j_{\mathrm{STJ}}[n]
 &=\exp\!\left\{\mathrm{j}\left(2\pi f_c t_n+\phi\right)\right\}, \\
 j_{\mathrm{MTJ}}[n]
 &=\frac{1}{\sqrt{M}}\sum_{m=1}^{M}\exp\!\left\{\mathrm{j}\left(2\pi f_m t_n+\phi_m\right)\right\}, \\
 j_{\mathrm{LFMJ}}[n]
 &=\exp\!\left\{\mathrm{j}2\pi\left(f_0t_n+\frac{\mu}{2}t_n^2\right)\right\}, \\
 j_{\mathrm{PTJ}}[n]
 &=b[n]\exp\!\left\{\mathrm{j}\left(2\pi f_p t_n+\phi_p\right)\right\}, \\
 j_{\mathrm{PBNJ}}[n]
 &=\left(\nu*h_{\mathrm{LP}}\right)[n]\exp\!\left(\mathrm{j}2\pi f_b t_n\right).
\end{align}
\end{subequations}
The discrete time is $t_n=n/F_s$. The parameters $f_c$ and $\phi$ denote the STJ carrier frequency and phase, while $M$, $f_m$, and $\phi_m$ describe the MTJ tones. The LFMJ parameters $f_0$ and $\mu$ are the starting frequency and chirp rate. The PTJ pulse mask $b[n]\in\{0,1\}$ is controlled by the repetition interval, pulse width, and optional timing jitter. For PBNJ, $\nu[n]$ is circular complex Gaussian noise, $h_{\mathrm{LP}}[n]$ is a low-pass shaping filter, and $f_b$ is the translated center frequency.

\subsection{Component-Set Representation and Mixed-Recognition Task}

\rev{For a record with active set $\mathcal{K}$, the target is a binary membership vector $\bm{y}\in\{0,1\}^{C}$, where $y_k=\mathbb{I}(k\in\mathcal{K})$. This is a \emph{multi-hot} vector: an entry equals one when its primitive is present and zero otherwise, and several entries may equal one simultaneously. The cardinality is $K=\sum_{k=1}^{C}y_k$. Here, $C=5$ is the number of available types, whereas $K$ is the number active in one record. Thus, the output identifies jamming types, not individual transmitters or separated waveforms. The measured singleton bank contains five classes, and the reported recognition task contains two-component and three-component mixtures. Simultaneous STJ and MTJ activation is excluded because their narrowband patterns can be ambiguous in the selected STFT representation, as in the compound-jamming setting of \cite{Jiang2026TCCN_ACSNet}. This is a task restriction, not a claim that the two signals cannot coexist physically.}

Let $\mathcal{A}_1$ denote the singleton masks and $\mathcal{A}_{\mathrm{mix}}$ denote the valid mixed masks with $K\in\{2,3\}$. Under the adopted restriction, $|\mathcal{A}_1|=5$ and $|\mathcal{A}_{\mathrm{mix}}|=16$. Singleton masks support primitive-evidence learning and an auxiliary singleton-versus-mixture task during training. Final inference is restricted to $\mathcal{A}_{\mathrm{mix}}$, so the model never returns a singleton in the reported mixed-interference test.

\subsection{Training-Listed and Held-Out Composition Split}

\rev{Let $\mathcal{A}_{\mathrm{list}}\subset\mathcal{A}_{\mathrm{mix}}$ contain the compositions admitted to auxiliary training, and let $\mathcal{A}_{\mathrm{hold}}=\mathcal{A}_{\mathrm{mix}}\setminus\mathcal{A}_{\mathrm{list}}$. The reported experiment uses ten training-listed and six held-out compositions, detailed in Table~\ref{tab:evaluation_partitions}. Every primitive in the held-out partition appears in the singleton training bank, but each complete held-out set is excluded from auxiliary training and model development. The 10/6 partition is an experimental choice, not a requirement of the network architecture. Consequently, the results demonstrate transfer on this particular partition of known jamming types; they do not establish invariance to the choice or number of held-out compositions. Unknown primitive types remain outside the task.}

\begin{figure*}[t!]
\centering
\includegraphics[width=0.98\textwidth]{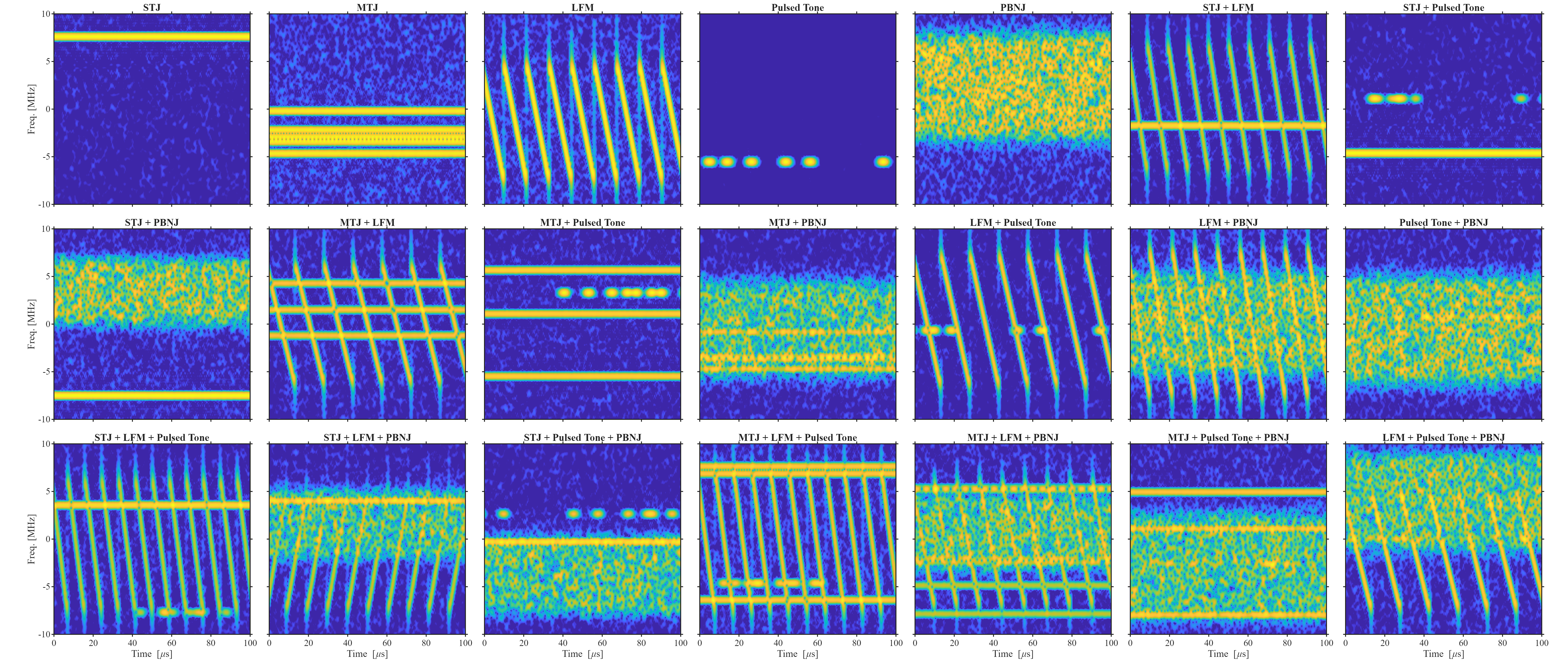}
\caption{Representative time-frequency representations of the five primitive jamming signals and their two-component and three-component compositions.}
\label{fig:overview}
\end{figure*}

\subsection{Singleton-to-Mixture Physical-Data Protocol}

\rev{The singleton-to-mixture data-use protocol separates three stages. First, gradient optimization uses measured singleton records together with auxiliary mixtures generated from their associated clean IQ; the auxiliary masks must belong to $\mathcal{A}_{\mathrm{list}}$. Second, a separate measured development set from $\mathcal{A}_{\mathrm{list}}$ supports checkpoint selection and decoder calibration. Third, after all model and decoder settings are fixed, independently acquired mixed test records from both $\mathcal{A}_{\mathrm{list}}$ and $\mathcal{A}_{\mathrm{hold}}$ are evaluated. Held-out compositions are excluded from early stopping, hyperparameter and threshold selection, and calibration as well as gradient training. Table~\ref{tab:protocol} shows which sample source can enter each stage.}

\begin{table*}[t]
\centering
\caption{Roles of Physical, Auxiliary, Development, and Test Samples}
\label{tab:protocol}
\setlength{\tabcolsep}{5pt}
\begin{tabular}{lcccc}
\toprule
Sample source & Gradient optimization & Model selection & Decoder calibration & Final test\\
\midrule
Measured singleton training records & Yes & No & No & No\\
Online mixed auxiliaries from $\mathcal{A}_{\mathrm{list}}$ & Yes & No & No & No\\
Measured development records from $\mathcal{A}_{\mathrm{list}}$ & No & Yes & Yes & No\\
Measured final-test records from $\mathcal{A}_{\mathrm{list}}$ & No & No & No & Yes\\
Measured final-test records from $\mathcal{A}_{\mathrm{hold}}$ & No & No & No & Yes\\
\bottomrule
\end{tabular}
\end{table*}

\rev{Thus, ``singleton-only physical training'' refers specifically to the measured records used for gradient updates. It does not mean that optimization uses only singleton labels, or that no measured mixtures are available for model development. Each optimization batch includes measured singletons and generated mixed examples, whereas the final recognition task is restricted to $K\in\{2,3\}$.}

\section{\rev{SPOC-Net Methodology}}
\label{sec:method}
\label{sec:analysis}

\begin{figure*}[t]
\centering
\includegraphics[width=0.98\textwidth]{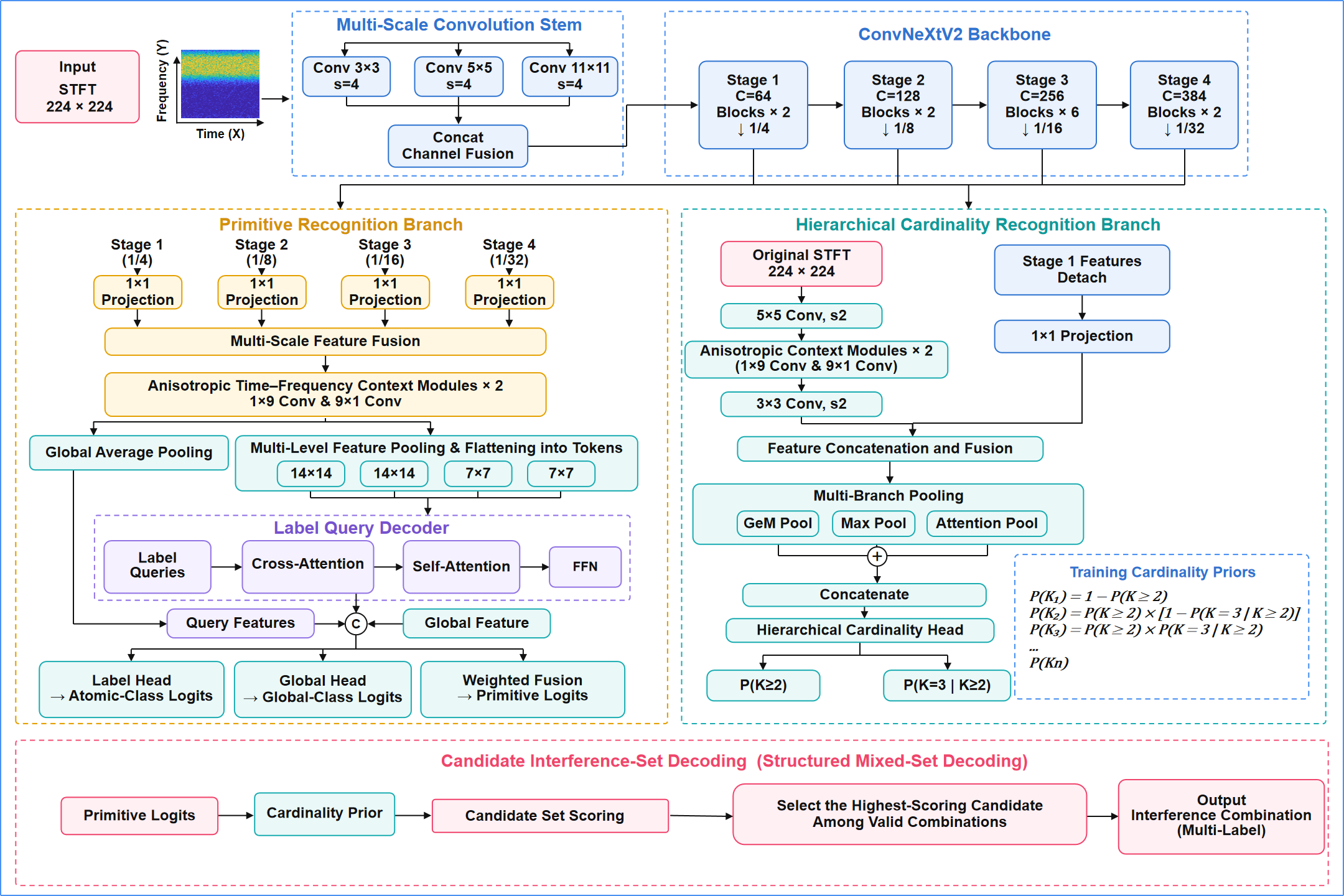}
\caption{Architecture of SPOC-Net. A shared multi-resolution spectro-temporal encoder feeds a primitive-evidence branch and a detached high-resolution cardinality branch. The singleton-versus-mixture output provides auxiliary training supervision. Final inference is restricted to the 16 valid two-component and three-component masks and uses the conditional two-versus-three cardinality probability.}
\label{fig:architecture}
\end{figure*}

\rev{SPOC-Net comprises a training-time signal composer and an image-based recognition network. This section presents the complete processing chain in that order: STFT representation, online generation of labeled mixtures, shared feature extraction, primitive and cardinality estimation, structured decoding, and the learning objective. Fig.~\ref{fig:architecture} summarizes the recognition architecture. During inference, only STFT preprocessing and the trained recognition network are used; the IQ composer is not run.}

\subsection{\rev{Time-Frequency Representation and Jamming Patterns}}
\label{sec:representation}

\rev{SPOC-Net uses a grayscale STFT image.} For a complex record $x[n]$, the transform and compressed logarithmic magnitude are
\begin{subequations}
\label{eq:stft_representation}
\begin{align}
 Z[m,\ell]
 &=\sum_n x[n]g[n-mR]\exp\!\left(-\mathrm{j}\frac{2\pi\ell n}{N_f}\right), \\
 D[m,\ell]
 &=20\log_{10}\!\left(\frac{|Z[m,\ell]|^{\gamma}}{\max_{m,\ell}|Z[m,\ell]|^{\gamma}}+\epsilon\right).
\end{align}
\end{subequations}
\rev{Here, $g[n]$ is a periodic Hann window, $R$ is the hop length, $N_f$ is the fast Fourier transform (FFT) size, $\gamma$ controls contrast, and $\epsilon$ prevents a logarithmic singularity. At the sampling frequency $F_s=20$ MHz specified in Section~\ref{sec:system}, the 128-sample window spans $6.4\,\mu$s, the 118-sample overlap spans $5.9\,\mu$s, and the 10-sample hop spans $0.5\,\mu$s. The transform uses $N_f=4096$. The Hann window and high overlap provide smooth time-frequency descriptions \cite{Cohen1989PROCIEEE_TFReview,Welch1967TAU_PSD}; zero-padding to the FFT size samples the spectrum on a finer frequency grid but does not increase the resolving power of the 128-sample window. With $\gamma=0.9$, the map is clipped to $[-35,0]$ dB, mapped to $[0,1]$, flipped vertically to preserve frequency orientation, resized bilinearly to $224\times224$, and stored as a single-channel image. The stored image is subsequently normalized to $[-1,1]$ for network input.}

\rev{Fig.~\ref{fig:overview} shows examples of the generated primitive signals and interference combinations whose STFT images are supplied to the recognizer. STJ produces a persistent horizontal line, MTJ produces several parallel lines, and LFMJ produces sloped or repeated chirp tracks. PTJ gives intermittent narrowband structures, whereas PBNJ appears as a band of stochastic texture. These are descriptions of the input signal patterns, not additional class definitions. In a mixture, unequal powers and spectral overlap can make one pattern much less visible than the others, motivating separate estimation of component identity and component count.}

\subsection{Online Clean-IQ Composition}
\label{sec:composer}

\rev{Let $s_i[n]$ denote the clean jammer IQ sequence associated with a measured singleton of type $i$ in the training bank. The composer uses these stored clean sequences rather than adding noisy receiver records. For a sampled two-component or three-component set in $\mathcal{A}_{\mathrm{list}}$, each sequence is centered and root-mean-square (RMS) normalized. Relative powers are drawn in decibels and centered across the selected components, yielding values $\rho_i$ and amplitude factors $a_i=10^{\rho_i/20}$. A fresh circular complex Gaussian background $v[n]$ is also centered and normalized. Denoting sample means by overbars, the resulting auxiliary record is}
\begin{subequations}
\label{eq:online_composition}
\begin{align}
 \widetilde{s}_i[n]
 &=\frac{s_i[n]-\overline{s}_i}{\operatorname{rms}(s_i-\overline{s}_i)}, \\
 u[n]
 &=\sum_{i\in\mathcal{K}}a_i\widetilde{s}_i[n], \\
 x_{\mathrm{aux}}[n]
 &=\sqrt{10^{\Gamma/10}}\frac{u[n]}{\operatorname{rms}(u)}
 +\frac{v[n]-\overline{v}}{\operatorname{rms}(v-\overline{v})}.
\end{align}
\end{subequations}
\rev{The target aggregate JNR is $\Gamma$. Relative component powers are sampled from $[-6,6]$ dB before centering, and $\Gamma$ is drawn from $\{-20,-15,-10,-5,0,5,10,15\}$ dB. Normalizing $u[n]$ sets the aggregate jammer power independently of the selected component count, while the normalized background has unit power. Each newly requested auxiliary example therefore has a known component mask and a controlled aggregate JNR. Composition precedes the STFT so that complex phases and overlapping spectra are combined before magnitude compression, rather than approximated by adding spectrogram images. The generated record then follows the same preprocessing as a measured record. This is a linear-superposition training model with a noise background, not a reproduction of every RF front-end effect.}

\rev{The composer also forms paired examples for cardinality learning: a two-component record and a three-component extension share the source sequences of their common primitives, the target aggregate JNR, and the background realization. Their component masks differ by one added primitive, with both masks restricted to $\mathcal{A}_{\mathrm{list}}$. Because the aggregate jammer signal is renormalized in each example, the received amplitudes of the shared components need not remain identical. The pair therefore supervises the effect of an additional component under fixed aggregate JNR, not an unconstrained increase in total jammer power.}

\subsection{Overall Architecture and Shared Spectro-Temporal Encoder}

\rev{Fig.~\ref{fig:architecture} details the network that processes measured and composed examples through the same image pipeline. Let $X\in[-1,1]^{1\times H\times W}$ be the normalized grayscale STFT image. The network produces primitive logits $\bm{z}\in\mathbb{R}^{C}$, an auxiliary singleton-versus-mixture logit, and a conditional two-versus-three cardinality logit. A logit is an unnormalized score that is converted to a probability by a sigmoid function. The training candidate set includes singletons and training-listed mixtures; the final decoder instead considers all protocol-valid mixed sets. These candidate sets and the final decision are}
\begin{subequations}
\label{eq:candidate_universes}
\begin{align}
 \mathcal{A}_{\mathrm{tr}}
 &=\mathcal{A}_1\cup\mathcal{A}_{\mathrm{list}}, \\
 \mathcal{A}_{\mathrm{eval}}
 &=\mathcal{A}_{\mathrm{mix}}, \\
 \widehat{\mathcal{K}}
 &=\arg\max_{A\in\mathcal{A}_{\mathrm{eval}}}\mathcal{S}_{\mathrm{mix}}(A\mid X).
\end{align}
\end{subequations}
The training universe contains the five singleton masks and the 10 masks admitted to online composition. The evaluation universe contains only the 16 valid two-component and three-component masks, so no singleton candidate is permitted in the reported test. Held-out compositions use the same five primitive logits and require no new output neuron.

The shared encoder augments the STFT with fixed frequency and time coordinates. For pixel indices $0\le i<H$ and $0\le j<W$, the coordinate maps and augmented input are
\begin{subequations}
\label{eq:coordinate_augmentation}
\begin{align}
 C_f(i,j)&=\frac{2i}{H-1}-1, \\
 C_t(i,j)&=\frac{2j}{W-1}-1, \\
 \overline{X}&=\operatorname{Concat}(X,C_f,C_t).
\end{align}
\end{subequations}
These maps retain absolute location information, which helps distinguish local structures with similar gradients but different positions or extents. A parallel stem observes the augmented input with several receptive fields. Let $\mathcal{R}$ be the kernel set, let $\Phi_l(\cdot)$ be encoder stage $l$, and let $D_l(\cdot)$ be stride-two downsampling. The encoder stages are
\begin{subequations}
\label{eq:shared_encoder}
\begin{align}
 F_1
 &=\Phi_1\!\left(\operatorname*{Concat}_{r\in\mathcal{R}}
 \left\{\operatorname{Conv}^{s=4}_{r\times r}(\overline{X})\right\}\right), \\
 F_l
 &=\Phi_l\!\left(D_l(F_{l-1})\right),\qquad l\in\{2,3,4\}.
\end{align}
\end{subequations}
The reported model uses $\mathcal{R}=\{3,5,11\}$, stage depths $(2,2,6,2)$, and channel dimensions $(64,128,256,384)$. The output scales are $1/4$, $1/8$, $1/16$, and $1/32$. The parallel stem preserves narrow ridges, short pulse fragments, and broad noise regions before deeper feature extraction.

Each encoder block uses a $7\times7$ depthwise convolution, channel normalization, pointwise expansion, a Gaussian error linear unit, global response normalization, pointwise projection, channel-spatial reweighting, and stochastic depth \cite{Liu2022CVPR_ConvNeXt,Woo2023CVPR_ConvNeXtV2,Woo2018ECCV_CBAM}. Features from all stages are projected to a common dimension $d$ and reduced to the spatial size of $F_4$ through adaptive average pooling (AAP). Their fusion and one anisotropic context unit are defined by
\begin{subequations}
\label{eq:multilevel_context}
\begin{align}
 U
 &=\mathcal{C}^{(2)}\!\left(\mathcal{N}\!\left[P_4(F_4)
 +\sum_{l=1}^{3}\operatorname{AAP}_{H_4\times W_4}\!\left(P_l(F_l)\right)\right]\right), \\
 \mathcal{C}(V)
 &=V+P_c\!\left(D_{1\times\kappa}\!\left(\mathcal{N}(V)\right)
 +D_{\kappa\times1}\!\left(\mathcal{N}(V)\right)\right).
\end{align}
\end{subequations}
Here, $P_l(\cdot)$ is a $1\times1$ projection, $\mathcal{N}(\cdot)$ is channel normalization, and $\mathcal{C}^{(2)}(\cdot)$ applies two context units. The implementation uses $\kappa=9$. The anisotropic depthwise kernels collect evidence mainly along time and frequency, which supports tones, pulse transitions, and chirp tracks with fewer parameters than a dense large two-dimensional kernel.

\subsection{Primitive-Evidence Modeling}

The primitive branch uses multi-level tokens together with a global average pooling (GAP) descriptor. Let $R_l=P_l(F_l)$ for $l\in\{1,2,3\}$ and let $R_4=U$. The token sequence and global descriptor are
\begin{subequations}
\label{eq:primitive_tokens}
\begin{align}
 T_l
 &=\operatorname{vec}\!\left(\operatorname{AAP}_{r_l\times r_l}(R_l)\right)
 +\bm{1}\bm{e}_l^{\mathsf T}, \\
 T
 &=\mathcal{N}_T\!\left(\operatorname{Concat}(T_1,T_2,T_3,T_4)\right), \\
 \bm{g}
 &=\operatorname{GAP}(U).
\end{align}
\end{subequations}
The learned vector $\bm{e}_l$ identifies the feature level, $\mathcal{N}_T$ is token normalization, and GAP produces $\bm{g}$. The token grids are $14\times14$, $14\times14$, $7\times7$, and $7\times7$. Shallow levels preserve local details, while deep levels provide compact semantic information.

One learnable query is assigned to each primitive type. A query-decoder layer applies cross-attention (CA), self-attention (SA), and a feed-forward network (FFN) through
\begin{subequations}
\label{eq:query_decoder}
\begin{align}
 \overline{Q}^{(m)}
 &=Q^{(m-1)}+\operatorname{CA}\!\left(Q^{(m-1)},T,T\right), \\
 \widetilde{Q}^{(m)}
 &=\overline{Q}^{(m)}+\operatorname{SA}\!\left(\overline{Q}^{(m)}\right), \\
 Q^{(m)}
 &=\widetilde{Q}^{(m)}+\operatorname{FFN}\!\left(\widetilde{Q}^{(m)}\right).
\end{align}
\end{subequations}
CA lets each query collect primitive-specific evidence from all feature levels, while SA models relations and competition among primitive types \cite{Vaswani2017NIPS_Attention,Liu2021Q2L,Ridnik2023WACV_MLDecoder}. The implementation uses two decoder layers.

Let $\bm{q}_k$ be the final query for primitive $k$. Query-conditioned evidence, global evidence, and their fusion are
\begin{subequations}
\label{eq:primitive_logits}
\begin{align}
 z_k^{(q)}
 &=h_q\!\left([\bm{q}_k;\bm{g}]\right), \\
 \bm{z}^{(g)}
 &=h_g\!\left(\bm{g}\right), \\
 \bm{z}
 &=\bm{z}^{(q)}+\alpha_g\bm{z}^{(g)}.
\end{align}
\end{subequations}
The query term emphasizes class-specific local shape, while the global term stabilizes decisions for primitives with strong full-image signatures. Table~\ref{tab:model_settings} gives the value of $\alpha_g$.

\subsection{Hierarchical Cardinality Training and Mixed-Set Decoding}

\rev{Primitive identity and set size require related but different evidence. Downsampling compresses the spatial grid of an STFT feature map. A weak tone ridge or short pulse fragment may then be averaged with surrounding background, so a feature indicating a third jamming type can become less distinct even when the two dominant types remain recognizable. This motivates estimating cardinality from a higher-resolution image path and the shallow feature $F_1$, rather than relying only on the deepest pooled descriptor. The image path, detached shallow path, and fused evidence are}
\begin{subequations}
\label{eq:cardinality_features}
\begin{align}
 E_x
 &=\Psi_x(X), \\
 E_s
 &=P_s\!\left(\operatorname{sg}(F_1)\right), \\
 E
 &=\Psi_f\!\left(\operatorname{Concat}\!\left(E_x,\operatorname{Resize}(E_s)\right)\right).
\end{align}
\end{subequations}
\rev{The stop-gradient operator $\operatorname{sg}(\cdot)$ passes feature values forward but blocks gradients through this connection, preventing this cardinality path from updating the shared encoder.} The image path contains a stride-two $5\times5$ convolution, two anisotropic context units, and a stride-two $3\times3$ convolution. The image path and cardinality predictor remain trainable.

High-resolution evidence is summarized through generalized-mean (GeM), max, and attention pooling. \rev{For channel $c$ and spatial feature vector $\bm{e}_j$, let $N_s$ be the number of spatial locations in $E$ and $p>0$ the GeM exponent. The pooled descriptors are}
\begin{subequations}
\label{eq:cardinality_pooling}
\begin{align}
 g_{\mathrm{gem},c}
 &=\left(\frac{1}{N_s}\sum_{j=1}^{N_s}|E_{c,j}|^p\right)^{1/p}, \\
 \bm{g}_{\max}
 &=\max_j\bm{e}_j, \\
 a_j
 &=\frac{\exp(\bm{w}^{\mathsf T}\bm{e}_j)}{\sum_n\exp(\bm{w}^{\mathsf T}\bm{e}_n)}, \\
 \bm{g}_{\mathrm{att}}
 &=\sum_{j=1}^{N_s}a_j\bm{e}_j, \\
 \bm{e}_{\mathrm{card}}
 &=[\bm{g}_{\mathrm{gem}};\bm{g}_{\max};\bm{g}_{\mathrm{att}}].
\end{align}
\end{subequations}
GeM pooling captures distributed energy, max pooling retains isolated peaks, and attention pooling focuses on informative regions \cite{Radenovic2019TPAMI_GeM}. Their combination helps when a weak third component occupies only a small part of the STFT.

\rev{The pooled evidence is combined with the global descriptor and class-order-invariant score summaries from the primitive branch as}
\begin{equation}
\label{eq:cardinality_descriptor}
\begin{split}
\bm{r}_{\mathrm{card}}=[&\bm{e}_{\mathrm{card}};
\operatorname{sg}(\bm{g});
\operatorname{sort}(\operatorname{sg}(\bm{z}));\\
&\operatorname{sort}(\operatorname{sigmoid}(\operatorname{sg}(\bm{z})))].
\end{split}
\end{equation}
The sorted logit and probability profiles describe how many primitives have strong support without tying the set-size decision to a fixed class identity. A shared predictor maps $\bm{r}_{\mathrm{card}}$ to an auxiliary logit $u_{\mathrm{mix}}$ for $K\ge2$ and a conditional logit $u_{3\mid\mathrm{mix}}$ for $K=3$ given a mixed observation. \rev{With cardinality temperature $T_c$, the predicted cardinality probabilities are}
\begin{subequations}
\label{eq:cardinality_probabilities}
\begin{align}
 \pi_{\mathrm{mix}}
 &=\operatorname{sigmoid}(u_{\mathrm{mix}}/T_c), \\
 \pi_{3\mid\mathrm{mix}}
 &=\operatorname{sigmoid}(u_{3\mid\mathrm{mix}}/T_c), \\
 p_1
 &=1-\pi_{\mathrm{mix}}, \\
 p_2
 &=\pi_{\mathrm{mix}}\left(1-\pi_{3\mid\mathrm{mix}}\right), \\
 p_3
 &=\pi_{\mathrm{mix}}\pi_{3\mid\mathrm{mix}}, \\
 \overline{p}_2
 &=1-\pi_{3\mid\mathrm{mix}}, \\
 \overline{p}_3
 &=\pi_{3\mid\mathrm{mix}}.
\end{align}
\end{subequations}
\rev{For each input image, $p_1$, $p_2$, and $p_3$ are nonnegative predicted probabilities for one, two, and three active types, respectively, and sum to one. These input-dependent probabilities act as cardinality priors when candidate sets are scored. Training uses all three because the training dataset includes singleton and mixed examples. The mixed-only test task uses the conditional probabilities $\overline{p}_2$ and $\overline{p}_3$, which sum to one over the two permitted test cardinalities. Thus, the auxiliary singleton-versus-mixture output is not used to rank final mixed candidates.}

\rev{Thresholding each primitive probability independently does not enforce the task constraints. For example, it can select both STJ and MTJ, select fewer than two or more than three types, or omit a weak third type whose score lies just below the threshold. SPOC-Net instead scores complete admissible candidates using both component evidence and the predicted set size. Let $\bm{a}\in\{0,1\}^{C}$ be the membership mask of candidate $A$. The primitive evidence, training score, mixed-test score, and their ranking equivalence for $A\in\mathcal{A}_{\mathrm{mix}}$ are}
\begin{subequations}
\label{eq:structured_scores}
\begin{align}
 \mathcal{B}(A\mid X)
 &=\sum_{k=1}^{C}a_k\log\operatorname{sigmoid}\!\left(\frac{z_k}{T_p}\right) \notag\\
 &\quad+\lambda_n\sum_{k=1}^{C}(1-a_k)
 \log\operatorname{sigmoid}\!\left(-\frac{z_k}{T_p}\right), \\
 \mathcal{S}_{\mathrm{tr}}(A\mid X)
 &=\mathcal{B}(A\mid X)+\lambda_c\log p_{|A|}
 +\frac{\beta_3}{T_c}\mathbb{I}(|A|=3), \\
 \mathcal{S}_{\mathrm{mix}}(A\mid X)
 &=\mathcal{B}(A\mid X)+\lambda_c\log\overline{p}_{|A|}
 +\frac{\beta_3}{T_c}\mathbb{I}(|A|=3), \\
 \mathcal{S}_{\mathrm{tr}}(A\mid X)
 &=\mathcal{S}_{\mathrm{mix}}(A\mid X)+\lambda_c\log\pi_{\mathrm{mix}}.
\end{align}
\end{subequations}
\rev{Here, $T_p$ is the primitive temperature, $\lambda_n$ weights evidence for absent components, $\lambda_c$ weights the cardinality probability, and $\beta_3$ adjusts the preference for three-component candidates. The component terms define a weighted Bernoulli log-score, reducing to the standard Bernoulli log-likelihood when $\lambda_n=1$. For every mixed candidate, $p_{|A|}=\pi_{\mathrm{mix}}\overline{p}_{|A|}$, so the term $\lambda_c\log\pi_{\mathrm{mix}}$ is common to all mixed candidates and cancels in their ranking.}

\rev{Decoder calibration uses the measured training-listed development records defined in Table~\ref{tab:protocol}. The candidate masks themselves are fixed by the recognition task and have no learned parameters. Let $\theta=(T_c,\lambda_c,\beta_3)$ and let $\Theta$ be the calibration grid. To avoid improving three-component recognition at an unacceptable cost to two-component recognition, calibration imposes a two-component exact-set accuracy floor $A_2^{\mathrm{floor}}$ with tolerance $\epsilon$. The feasible set and selection rule are}
\begin{subequations}
\label{eq:decoder_calibration}
\begin{align}
 \Theta_{\mathrm{feas}}
 &=\left\{\theta\in\Theta:\operatorname{Acc}_{2}^{\mathrm{dev}}(\theta)
 \ge A_2^{\mathrm{floor}}-\epsilon\right\}, \\
 \theta^{\star}
 &=\underset{\theta\in\Theta_{\mathrm{feas}}}{\operatorname{lexmax}}
 \left[\operatorname{Acc}_{3}^{\mathrm{dev}}(\theta),
 \operatorname{BalAcc}_{2,3}^{\mathrm{dev}}(\theta),
 \operatorname{Acc}_{\mathrm{all}}^{\mathrm{dev}}(\theta)\right].
\end{align}
\end{subequations}
\rev{Here, $\operatorname{Acc}_{r}^{\mathrm{dev}}$ denotes exact-set accuracy for cardinality $r$, $\operatorname{BalAcc}_{2,3}^{\mathrm{dev}}$ averages the two cardinality-specific accuracies, and $\operatorname{Acc}_{\mathrm{all}}^{\mathrm{dev}}$ pools the development records. The lexicographic rule first maximizes three-component accuracy, then uses balanced accuracy and overall accuracy to break ties. The checkpoint and selected decoder settings are frozen before final testing.}

\subsection{Learning Objective and Implementation Details}

The learning objective covers primitive membership, valid-set discrimination, auxiliary singleton-versus-mixture separation, conditional two-versus-three recognition, and the response to adding a third component. Let $p_{ik}=\operatorname{sigmoid}(z_{ik})$ for sample $i$ and primitive $k$. The shifted negative probability, one-label ASL term, and cardinality-balanced classification loss are
\begin{subequations}
\label{eq:asymmetric_classification_loss}
\begin{align}
 \overline{p}_{ik}^{-}
 &=\min\!\left(1,1-p_{ik}+m_a\right), \\
 \ell_{ik}^{\mathrm{asy}}
 &=-y_{ik}(1-p_{ik})^{\gamma_{+}}\log p_{ik} \notag\\
 &\quad-(1-y_{ik})\left(1-\overline{p}_{ik}^{-}\right)^{\gamma_{-}}
 \log\overline{p}_{ik}^{-}, \\
 \mathcal{L}_{\mathrm{cls}}
 &=\frac{1}{|\mathcal{R}_{\mathcal{B}}|}
 \sum_{r\in\mathcal{R}_{\mathcal{B}}}
 \frac{\sum_{i\in\mathcal{B}_r}\sum_{k=1}^{C}\ell_{ik}^{\mathrm{asy}}}{C|\mathcal{B}_r|}.
\end{align}
\end{subequations}
\rev{Let $B$ denote the batch size. Here, $\mathcal{B}_r=\{i:|\mathcal{K}_i|=r\}$} and $\mathcal{R}_{\mathcal{B}}=\{r:|\mathcal{B}_r|>0\}$. ASL reduces the influence of easy negative labels while retaining hard negative evidence \cite{Ridnik2021ICCV_ASL}, and the outer average limits imbalance among set sizes.

The candidate-set loss uses $\mathcal{S}_{\mathrm{tr}}$ and the training universe $\mathcal{A}_{\mathrm{tr}}$. For ground-truth set $A_i$, it is
\begin{equation}
\label{eq:set_loss}
\mathcal{L}_{\mathrm{set}}
=-\frac{1}{B}\sum_{i=1}^{B}
\log\frac{\exp\mathcal{S}_{\mathrm{tr}}(A_i\mid X_i)}
{\sum_{A\in\mathcal{A}_{\mathrm{tr}}}\exp\mathcal{S}_{\mathrm{tr}}(A\mid X_i)}.
\end{equation}
Measured singleton samples and online-composed mixed samples compete against the five singleton masks and the 10 training-listed mixed masks. Held-out compositions are absent from this loss.

Let $\ell_{\mathrm{bce}}(u,t)$ denote binary cross-entropy (BCE) with logits, and let $\mathcal{B}_{\mathrm{mix}}=\{i:|\mathcal{K}_i|\ge2\}$. The auxiliary, conditional, and combined cardinality losses are
\begin{subequations}
\label{eq:cardinality_losses}
\begin{align}
 \mathcal{L}_{\mathrm{mix}}
 &=\frac{1}{B}\sum_{i=1}^{B}\ell_{\mathrm{bce}}\!\left(
 u_{\mathrm{mix},i},\mathbb{I}(|\mathcal{K}_i|\ge2)\right), \\
 \mathcal{L}_{3\mid\mathrm{mix}}
 &=\frac{1}{|\mathcal{B}_{\mathrm{mix}}|}
 \sum_{i\in\mathcal{B}_{\mathrm{mix}}}w_{|\mathcal{K}_i|}
 \ell_{\mathrm{bce}}\!\left(u_{3\mid\mathrm{mix},i},
 \mathbb{I}(|\mathcal{K}_i|=3)\right), \\
 \mathcal{L}_{\mathrm{card}}
 &=\frac{1}{2}\left(\mathcal{L}_{\mathrm{mix}}+\mathcal{L}_{3\mid\mathrm{mix}}\right).
\end{align}
\end{subequations}
The auxiliary loss separates singleton and mixed training samples, while the conditional loss directly controls the two-versus-three decision used during final inference. The decoder uses only $\pi_{3\mid\mathrm{mix}}$ through $\overline{p}_2$ and $\overline{p}_3$.

\rev{For the paired examples defined in Section~\ref{sec:composer}, let $u_{3\mid\mathrm{mix},p}^{(2)}$ and $u_{3\mid\mathrm{mix},p}^{(3)}$ be the conditional three-component logits of the two-component and three-component records, respectively. With $N_p$ pairs and margin $m_p$, the monotonic pair loss and complete objective are}
\begin{subequations}
\label{eq:pair_and_total_loss}
\begin{align}
 \mathcal{L}_{\mathrm{pair}}
 &=\frac{1}{N_p}\sum_{p=1}^{N_p}\max\!\left(
 0,m_p-u_{3\mid\mathrm{mix},p}^{(3)}+u_{3\mid\mathrm{mix},p}^{(2)}\right), \\
 \mathcal{L}
 &=\lambda_{\mathrm{cls}}\mathcal{L}_{\mathrm{cls}}
 +\lambda_{\mathrm{set}}\mathcal{L}_{\mathrm{set}}
 +\lambda_{\mathrm{card}}\mathcal{L}_{\mathrm{card}}
 +\lambda_{\mathrm{pair}}\mathcal{L}_{\mathrm{pair}}.
\end{align}
\end{subequations}
\rev{The paired term encourages stronger three-component evidence for the extension than for its two-component counterpart. Each batch is approximately balanced among one-, two-, and three-component examples, with measured singletons and online-composed mixtures used according to Table~\ref{tab:protocol}.}

AdamW is used for optimization \cite{Loshchilov2019ICLR_AdamW}. Dropout and stochastic depth limit overfitting \cite{Srivastava2014JMLR_Dropout}, and the learning rate follows cosine annealing with warm restarts \cite{Loshchilov2017ICLR_SGDR}. Mixed-precision training, gradient clipping, exponential moving average (EMA) weights, and light time-frequency perturbations are also used. The default model has about 14.05 million trainable parameters. Tables~\ref{tab:model_settings} and \ref{tab:training_settings} list the architecture and training settings.

\begin{table}[t]
\centering
\caption{Architecture and Decoder Configuration}
\label{tab:model_settings}
\setlength{\tabcolsep}{4pt}
\begin{tabular}{ll}
\toprule
Setting & Value\\
\midrule
Network input & $224\times224$ grayscale STFT only\\
Coordinate augmentation & Time and frequency maps\\
Stem kernels and stride & $3,5,11$ and $4$\\
Stage depths & $(2,2,6,2)$\\
Stage channels & $(64,128,256,384)$\\
Decoder dimension $d$ & $384$\\
Token grids & $14,14,7,7$\\
Query layers and heads & $2$ and $6$\\
Primitive context units & $2$\\
Cardinality context units & $2$ before fusion and $1$ after fusion\\
Global-logit coefficient $\alpha_g$ & $0.35$\\
Cardinality evidence channels & $64$\\
Cardinality hidden dimension & $256$\\
JNR or metadata side input & None\\
Auxiliary cardinality output & Singleton versus mixture\\
Final cardinality output & Two versus three components\\
Final candidate universe & 16 mixed masks\\
Dropout and stochastic depth & $0.15$ and $0.10$\\
Trainable parameters & $14.05$ million\\
\bottomrule
\end{tabular}
\end{table}

\begin{table*}[t]
\centering
\caption{Training, Auxiliary Composition, Loss, Protocol, and Calibration Settings}
\label{tab:training_settings}
\setlength{\tabcolsep}{4pt}
\footnotesize
\begin{tabularx}{\textwidth}{p{0.15\textwidth}X p{0.15\textwidth}X}
\toprule
Category & Setting & Category & Setting\\
\midrule
Optimization
& 120 epochs; batch size 32
& Optimizer
& AdamW\\
Learning rate
& $2.5\times10^{-4}$ to $10^{-6}$
& Weight decay
& $7\times10^{-4}$\\
Gradient handling
& Norm clipping at $1.0$
& EMA decay
& $0.999$\\
Physical gradient data
& Measured singleton records only
& Auxiliary source
& Clean-IQ singleton bank only\\
Training mixed masks
& $\mathcal{A}_{\mathrm{list}}$ only
& Training candidate universe
& $\mathcal{A}_1\cup\mathcal{A}_{\mathrm{list}}$\\
Development compositions
& $\mathcal{A}_{\mathrm{list}}$ only
& Held-out development records
& None\\
Final candidate cardinalities
& $K\in\{2,3\}$
& Singleton-versus-mixture output
& Auxiliary training only\\
Batch cardinalities
& Approximately balanced one, two, and three components
& Paired auxiliaries
& 4 pairs per batch\\
Relative component power
& $[-6,6]$ dB
& Auxiliary JNR grid
& $-20{:}5{:}15$ dB\\
Low-JNR sampling
& $1.3\times$ for JNR $\leq-10$ dB
& Very-low-JNR sampling
& Additional $1.8\times$ for JNR $\leq-20$ dB\\
Loss weights
& $1.00,\ 0.25,\ 0.50,\ 0.15$
& ASL $(\gamma_+,\gamma_-)$
& $(0,4)$\\
ASL negative shift $m_a$
& $0.05$
& Pair margin $m_p$
& $0.50$\\
Conditional costs $(w_2,w_3)$
& $(1.50,1.25)$
& View averaging
& 2 for calibration and selection; 4 for test\\
Calibration records
& Measured $\mathcal{A}_{\mathrm{list}}$ development set
& Calibration score
& $\mathcal{S}_{\mathrm{mix}}$\\
Calibration $T_c$
& $\{0.70,0.85,1.00,1.20,1.40\}$
& Calibration $\lambda_c$
& $\{0.50,0.75,1.00,1.25,1.50,2.00\}$\\
Calibration $\beta_3$
& $-1.00{:}0.25{:}1.25$
& Two-component floor tolerance $\epsilon$
& $0.005$\\
\bottomrule
\end{tabularx}
\end{table*}

\section{Experimental Setup}
\label{sec:experiments}

\subsection{Computing Environment and Conducted RF Acquisition Platform}

Training and inference are carried out on a workstation with an Intel Core i9-12900KF processor, 64 GB of system memory, and an NVIDIA GeForce RTX 5060 Ti with 16 GB of memory. The implementation uses Python 3.11 and PyTorch 2.x with NVIDIA Compute Unified Device Architecture (CUDA) and the CUDA Deep Neural Network library (cuDNN) for acceleration.

\begin{figure*}[!t]
\centering
\begin{minipage}[t]{0.48\textwidth}
    \centering
    \includegraphics[width=\linewidth]{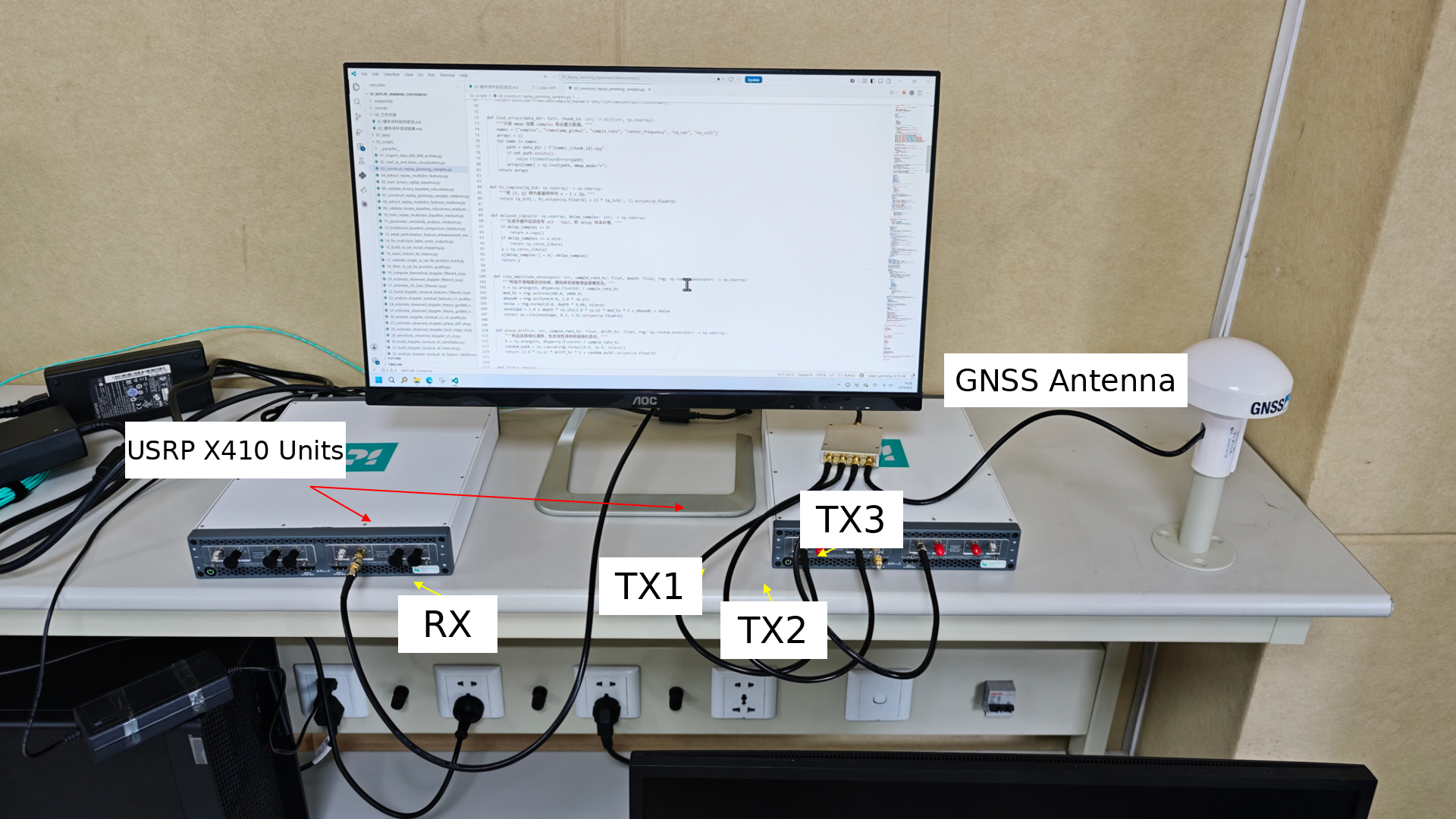}\\[0.6ex]
    (a)
\end{minipage}
\hfill
\begin{minipage}[t]{0.48\textwidth}
    \centering
    \includegraphics[width=\linewidth]{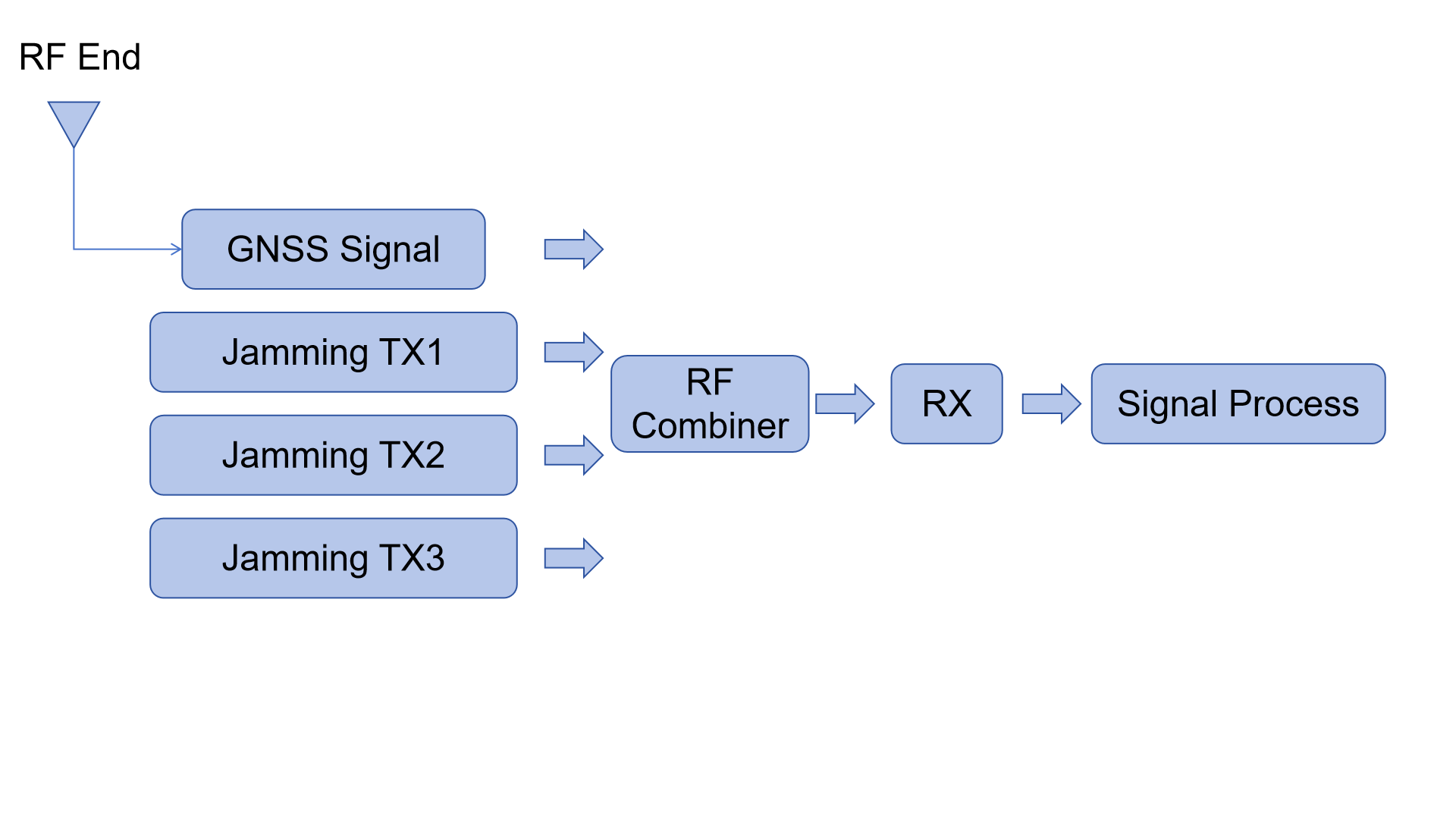}\\[0.6ex]
    (b)
\end{minipage}
\caption{Conducted RF acquisition platform. Panel (a) shows the hardware implementation with two USRP X410 units, a GNSS antenna, and a four-input RF combiner. Panel (b) shows the signal flow: the live GNSS antenna signal and up to three jammer transmit channels are combined in the RF domain, recorded by the USRP receive channel, and passed to the IQ-processing pipeline.}
\label{fig:physical_platform}
\end{figure*}

The dataset is collected with the conducted RF acquisition platform shown in Fig.~\ref{fig:physical_platform}. An ANTF004-SMAJ GNSS antenna receives live satellite signals and feeds one input of a four-input RF combiner through a coaxial cable. One Universal Software Radio Peripheral (USRP) X410 is configured with up to three independently controlled transmit channels, denoted TX1--TX3, \rev{to generate the selected jamming signals.} Each transmit channel is connected conductively to a separate combiner input. The combiner output is delivered by coaxial cable to the receive channel of a second USRP X410, which records the complex IQ sequence for subsequent signal processing. \rev{Each acquired record therefore contains the live GNSS signal, receiver noise, and the selected one-, two-, or three-component jamming set.}

\rev{Each processed signal snapshot is one $1$-ms record sampled at $F_s=20$ MHz, containing $N=20{,}000$ complex samples as defined in Section~\ref{sec:system}. The component set, cardinality, JNR label, and IQ sequence are stored for every record. The jammer transmit channels are activated for the target set, and their output levels are controlled independently to achieve the specified component-power relations and aggregate JNR. Singleton acquisition provides the measured receiver records and associated clean jammer IQ used by the composer. Measured mixtures are produced by simultaneous conducted RF injection and recorded independently at the X410 acquisition receiver. Their uses in development, comparison-model training, and testing are distinguished below. This use of physical measurements is related to prior work on measured GNSS interference and practical recognition \cite{Ott2024ICLGNSS_FewShot,Spanghero2025TIFS_JammerLocalization,Jia2025TIM_LowPowerCNN,Xiao2025TAES_CompoundFusion}.}

\subsection{Evaluation Protocol and Composition Partitions}

\rev{The final test set contains $N_{\mathrm{rec}}=14{,}220$ independently acquired $1$-ms conducted mixed records, covering the 16 valid compositions and JNR values from $-20$ to $15$ dB in 5-dB steps. Here and in the result tables, $N_{\mathrm{rec}}$ counts records in the indicated subset; $N$ remains the number of IQ samples per record. The training-listed test subset contains 8,887 records from the ten compositions in $\mathcal{A}_{\mathrm{list}}$, and the held-out test subset contains 5,333 records from the six compositions in $\mathcal{A}_{\mathrm{hold}}$.}

\rev{The measured development set contains only $\mathcal{A}_{\mathrm{list}}$ compositions and is disjoint from both final-test subsets. It supports checkpoint selection, early stopping when used, hyperparameter selection, and decoder calibration. No waveform, image, label, or aggregate result from $\mathcal{A}_{\mathrm{hold}}$ is used for these decisions. After development, each final-test partition is evaluated with the same fixed network and decoder, using the view-averaging settings reported for that model. The same composition-exclusion rule applies to the comparison models; their training and output protocols are specified in Section~\ref{sec:comparison_protocol}.}

\begin{table*}[t]
\centering
\caption{Composition-Level Final-Test Partitions}
\label{tab:evaluation_partitions}
\setlength{\tabcolsep}{4pt}
\begin{tabularx}{\textwidth}{lccX}
\toprule
Partition & Classes & \rev{$N_{\mathrm{rec}}$} & Component sets\\
\midrule
Training-listed final test & 10 & 8,887 & STJ+LFMJ, STJ+PBNJ, MTJ+LFMJ, MTJ+PTJ, LFMJ+PTJ, PTJ+PBNJ, STJ+LFMJ+PBNJ, STJ+PTJ+PBNJ, MTJ+LFMJ+PTJ, MTJ+LFMJ+PBNJ\\
Held-out final test & 6 & 5,333 & STJ+PTJ, MTJ+PBNJ, LFMJ+PBNJ, STJ+LFMJ+PTJ, MTJ+PTJ+PBNJ, LFMJ+PTJ+PBNJ\\
Full mixed final test & 16 & 14,220 & Union of the training-listed and held-out final-test partitions\\
\bottomrule
\end{tabularx}
\end{table*}

\subsection{Comparison Models and Component-Set Output Protocol}
\label{sec:comparison_protocol}

\rev{We compare SPOC-Net with six reference architectures for mixed-jamming component-set recognition: AlexNet \cite{Krizhevsky2012NIPS_AlexNet}, ResNet18 \cite{He2016CVPR_ResNet}, Feature-ResNet18, ACSNet \cite{Jiang2026TCCN_ACSNet}, MSFF-KAN, and the Temporal-Spatial Feature Aggregation Network (TSFANet) \cite{Jia2025TAES_MSFFKAN,Zhong2024TIM_TSFANet}. Feature-ResNet18 is a ResNet18-based reference augmented with signal statistics, as specified below. The reference architectures are adapted to a common five-label primitive-set output and are denoted by the suffix ``-ML.'' This makes their output space comparable with SPOC-Net and allows evaluation on new combinations of known labels, unlike a fixed-list compound-class softmax head.}

The adapted final layer contains one logit for each primitive in the order STJ, MTJ, LFMJ, PTJ, and PBNJ. The five logits are trained with ASL and converted to sigmoid probabilities. A composition uses a multi-hot primitive mask, so STJ+PTJ is represented by $[1,0,0,1,0]$. This shared output form allows every architecture to return a held-out composition from known primitive evidence. \rev{A strict 10-class softmax head cannot predict an unlisted composition and is not used for the component-set comparisons reported here.}

\rev{Index the six adapted reference architectures above by $b\in\{1,\ldots,6\}$. For architecture $b$, let $z_{b,k}$ be the logit for primitive $k$, $T_b$ its calibration temperature, $\delta_{b,k}$ a component-specific bias, and $\bm{a}$ the membership mask of candidate $A$. Each reference model uses the following calibrated probability, candidate score, and final decision:}
\begin{subequations}
\label{eq:baseline_decoder}
\begin{align}
 p_{b,k}
 &=\operatorname{sigmoid}\!\left(\frac{z_{b,k}+\delta_{b,k}}{T_b}\right), \\
 \mathcal{S}_b(A)
 &=\sum_{k=1}^{5}a_k\log p_{b,k} \notag\\
 &\quad+\lambda_b\sum_{k=1}^{5}(1-a_k)\log(1-p_{b,k}) \notag\\
 &\quad+\beta_b\mathbb{I}(|A|=3), \\
 \widehat{\mathcal{K}}_b
 &=\arg\max_{A\in\mathcal{A}_{\mathrm{mix}}}\mathcal{S}_b(A).
\end{align}
\end{subequations}
The fixed set $\mathcal{A}_{\mathrm{mix}}$ contains all 16 protocol-valid two-component and three-component masks, including the six held-out masks. These masks follow the task definition and contain no learned information. The parameters $T_b$, $\lambda_b$, $\beta_b$, and $\delta_{b,k}$ are selected only on training-listed development data. The comparison decoder has no label-query module, detached cardinality branch, paired loss, or set-level training loss. 

\rev{The reference models receive measured singleton records and measured mixtures from the ten training-listed compositions for gradient training, with no online IQ or image composition. SPOC-Net instead uses measured singletons and online-composed mixtures, as in Table~\ref{tab:protocol}. Both protocols exclude $\mathcal{A}_{\mathrm{hold}}$ compositions from model development. Thus, the reference models provide a measured-mixture-supervised benchmark, rather than a matched-data ablation of the proposed architecture. Their training-listed results assess recognition with direct mixture supervision, whereas their held-out results assess transfer beyond those combinations.}

All comparison models use $224\times224$ grayscale STFT inputs and random initialization. ImageNet pretraining is not used. They are trained with AdamW for 120 epochs, batch size 32, a common balanced sampler, and the same baseline augmentation policy. The augmentation includes time and frequency shifts, time and frequency flips, intensity and bias changes, light image noise, random erasing, and stripe masking. EMA weights with decay $0.999$ are used for evaluation. \rev{Each reference model uses one test view per record, whereas SPOC-Net uses four-view averaging (Table~\ref{tab:training_settings}). This difference is retained in the reported results; the comparison therefore reflects the complete inference procedures as well as the different training data.} Decoder calibration uses only training-listed development data and never uses a held-out waveform or label.

AlexNet-ML and ResNet18-ML use their canonical convolutional backbones with a five-output final layer. ACSNet-ML keeps its asymmetric convolution blocks and replaces the original closed-set classifier with the five-output head. MSFF-KAN-ML and TSFANet-ML are our implementations of the published architectures, adapted to five-label prediction. Feature-ResNet18-ML combines the 512-dimensional ResNet18 image descriptor with 14 signal statistics. The statistics contain 11 time-domain values and three frequency-domain values. For the reported experiments, the 14 signal statistics were computed from the recorded IQ sequences. The 14 values are mapped to 128 dimensions, concatenated with the image descriptor, and passed through a 512-dimensional fusion layer before the five-output head.

\begin{table*}[t]
\centering
\caption{Training and Inference Protocol for the Component-Set Comparison Models}
\label{tab:comparison_protocol}
\setlength{\tabcolsep}{4.0pt}
\begin{tabularx}{\textwidth}{p{0.145\textwidth}X p{0.145\textwidth}X}
\toprule
Category & Setting & Category & Setting\\
\midrule
Prediction task
& Five-label primitive-set prediction
& Output and loss
& Five logits, sigmoid, and ASL\\
Gradient data
& Measured singletons and measured $\mathcal{A}_{\mathrm{list}}$ mixtures
& Online composition
& None\\
Held-out development data
& None
& Final candidate universe
& All 16 valid two-component and three-component masks\\
Primary image input
& $224\times224$ grayscale STFT
& JNR side information
& None for every model\\
Initialization
& Random, without ImageNet pretraining
& Optimization
& AdamW, 120 epochs, batch size 32\\
Augmentation
& Shift, flip, jitter, noise, erase, and stripe mask
& Sampling
& Composition-balanced with higher three-component and low-JNR weight\\
EMA
& Decay $0.999$
& Test-time averaging
& One view\\
Calibration
& Training-listed development data only
& Calibrated terms
& Temperature, negative weight, three-component bias, and component biases\\
Feature-ResNet18-ML
& ResNet18 plus 14 signal statistics
& \rev{Strict softmax heads}
& \rev{Not used in the reported component-set comparisons}\\
\bottomrule
\end{tabularx}
\end{table*}

\rev{The reference-model names in the figure legends omit the ``-ML'' suffix but refer throughout to the five-label versions specified here. None of those plotted component-set results uses a strict 10-class softmax classifier.}

\subsection{Evaluation Metrics}

\rev{For an evaluation subset containing $N_{\mathrm{rec}}$ records, let $\widehat{\bm{y}}_i$ and $\bm{y}_i$ be the predicted and true multi-hot membership vectors. Exact-set accuracy, Hamming loss, and cardinality accuracy are}
\begin{subequations}
\label{eq:evaluation_metrics}
\begin{align}
 \mathrm{Exact}
 &=\frac{1}{\rev{N_{\mathrm{rec}}}}\sum_{i=1}^{\rev{N_{\mathrm{rec}}}}\mathbb{I}(\widehat{\bm{y}}_i=\bm{y}_i), \\
 \mathrm{Hamming}
 &=\frac{1}{5\rev{N_{\mathrm{rec}}}}\sum_{i=1}^{\rev{N_{\mathrm{rec}}}}\sum_{k=1}^{5}
 \mathbb{I}(\widehat{y}_{ik}\neq y_{ik}), \\
 \mathrm{CardAcc}
 &=\frac{1}{\rev{N_{\mathrm{rec}}}}\sum_{i=1}^{\rev{N_{\mathrm{rec}}}}
 \mathbb{I}(|\widehat{\mathcal{K}}_i|=|\mathcal{K}_i|).
\end{align}
\end{subequations}
Micro-precision, micro-recall, and micro-F1 pool true positives, false positives, and false negatives over all records and primitive labels. Exact-set accuracy is the primary metric because it requires both correct primitive identities and correct set size.

\section{Results and Analysis}
\label{sec:results}

\rev{The results address three questions: how reliably SPOC-Net recovers complete component sets across JNR, whether this ability transfers to compositions excluded from model development, and which errors limit that transfer. We first examine aggregate and JNR-dependent performance, then compare the training-listed and held-out partitions, and finally relate composition-level errors to primitive recognition and cardinality estimation. All numerical results below use the test records and model protocols of Section~\ref{sec:experiments}.}

\subsection{Overall and JNR-Dependent Performance}

\rev{The aggregate results in Table~\ref{tab:overall_results} distinguish recovering most components from recovering the complete set. Exact-set accuracy is 80.69\%, whereas micro-F1 is 92.84\%: many unsuccessful set decisions therefore still contain correct primitive labels. Micro-precision exceeds micro-recall, indicating a greater tendency to omit active components than to introduce false ones. This matters for interference monitoring because identifying only the dominant jammer can leave an additional type unaccounted for. The difference between two-component and three-component exact-set accuracy, 88.35\% versus 70.83\%, locates the main difficulty in the more crowded mixtures rather than in component-set recognition uniformly.}

\begin{table*}[t]
\centering
\caption{Overall SPOC-Net Results on the Full Measured Mixed Set}
\label{tab:overall_results}
\setlength{\tabcolsep}{2pt}
\scriptsize
\begin{tabular}{lrrrrrrr}
\toprule
Test data & \rev{$N_{\mathrm{rec}}$} & \shortstack{Exact-set\\(\%)} & \shortstack{Micro-precision\\(\%)} & \shortstack{Micro-recall\\(\%)} & \shortstack{Micro-F1\\(\%)} & \shortstack{Cardinality accuracy\\(\%)} & \shortstack{Hamming loss\\(\%)}\\
\midrule
Full mixed set & 14,220 & 80.69 & 95.31 & 90.49 & 92.84 & 87.08 & 6.81\\
\bottomrule
\end{tabular}
\end{table*}

\rev{The JNR dependence in Table~\ref{tab:jnr_results} shows where this limitation is most severe. At nonnegative JNR, exact-set accuracy is at least 99.49\%; reducing JNR therefore exposes a detectability problem that is largely hidden by the high-JNR results. At $-15$ dB, two-component accuracy remains 82.65\%, but three-component accuracy falls to 11.60\%. At $-5$ dB, the corresponding values have recovered to 99.40\% and 91.93\%. Because JNR describes aggregate jammer power, it does not guarantee that every constituent is equally visible, especially when relative component powers are unequal. The pattern is consistent with a weak additional component being obscured by noise or overlap. These aggregate results alone cannot isolate the respective effects of STFT clipping, image resizing, and learned feature extraction.}

\begin{table*}[t]
\centering
\caption{SPOC-Net Performance Versus JNR on the Full Mixed Set}
\label{tab:jnr_results}
\begin{tabular}{crrrrr}
\toprule
JNR (dB) & \rev{$N_{\mathrm{rec}}$} & Exact-set (\%) & Micro-F1 (\%) & Two-component exact-set (\%) & Three-component exact-set (\%)\\
\midrule
$-20$ & 1,778 & 16.59 & 58.85 & 29.28 & 0.13\\
$-15$ & 1,773 & 51.55 & 85.56 & 82.65 & 11.60\\
$-10$ & 1,777 & 81.99 & 95.84 & 96.10 & 63.79\\
$-5$  & 1,776 & 96.11 & 99.18 & 99.40 & 91.93\\
0     & 1,778 & 99.49 & 99.90 & 99.70 & 99.23\\
5     & 1,777 & 99.72 & 99.94 & 99.90 & 99.49\\
10    & 1,785 & 99.94 & 99.98 & 100.00 & 99.87\\
15    & 1,776 & 100.00 & 100.00 & 100.00 & 100.00\\
\bottomrule
\end{tabular}
\end{table*}

\subsection{Generalization Across Composition Partitions}

\rev{Table~\ref{tab:split_comparison} separates two distinct questions: recognition of compositions represented during development and recognition of combinations never used for development. These are complementary views of the same final test set, not results from separately fitted models. The distinction is important because strong accuracy on the training-listed partition does not by itself demonstrate compositional generalization.}

\rev{On training-listed compositions, SPOC-Net is competitive but not the best-performing method: its 80.57\% exact-set accuracy is 1.43 percentage points below MSFF-KAN-ML. Direct measured-mixture supervision remains useful in this setting. The ranking changes on held-out combinations, where SPOC-Net achieves 80.89\%, compared with 62.12\% for ResNet18-ML, the strongest reference. Thus, the principal benefit is not a uniform improvement on familiar mixtures, but retention of recognition accuracy when known primitive types appear in an untrained combination.}

\rev{The similar training-listed and held-out averages for SPOC-Net should not be interpreted as evidence that every composition is equally difficult: the two partitions contain different component sets. Rather, together with the class-level results below, they support transfer on the specified split. Its leading full-set accuracy is driven by this held-out performance. Because the reference methods use measured mixed training data and one test view while SPOC-Net uses generated mixtures and four test views, the 18.77-point held-out margin measures the difference between complete reported procedures. It does not isolate the contribution of the composer, an individual network branch, or test-time averaging.}

\begin{table*}[t]
\centering
\caption{Exact-Set Accuracy of Five-Label Component-Set Models Across Composition Partitions}
\label{tab:split_comparison}
\setlength{\tabcolsep}{9pt}
\begin{tabular}{lccc}
\toprule
Model & Training-listed 10 classes (\%) & Held-out 6 classes (\%) & Full 16 classes (\%)\\
\midrule
AlexNet-ML & 75.48 & 34.97 & 60.29\\
ResNet18-ML & 79.00 & 62.12 & 72.67\\
Feature-ResNet18-ML & 80.45 & 59.48 & 72.59\\
ACSNet-ML & 78.87 & 43.95 & 65.77\\
MSFF-KAN-ML & \textbf{82.00} & 50.85 & 70.32\\
TSFANet-ML & 81.69 & 41.83 & 66.74\\
SPOC-Net & 80.57 & \textbf{80.89} & \textbf{80.69}\\
\bottomrule
\end{tabular}
\end{table*}

\begin{figure*}[t]
\centering
\includegraphics[width=\textwidth]{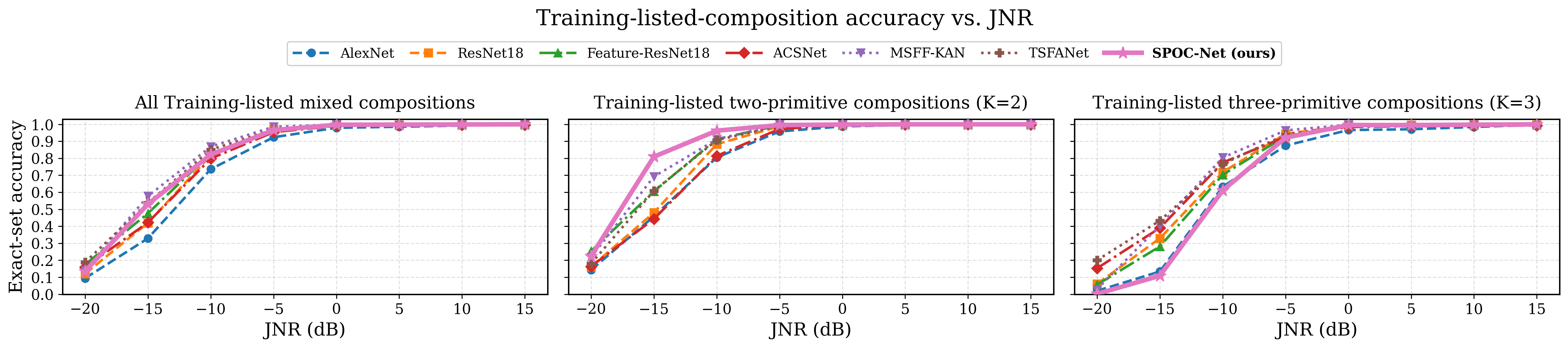}
\caption{Exact-set accuracy versus JNR on the 10 training-listed mixed compositions. The panels report overall, two-component, and three-component results for 8,887 measured records. The figure legend omits the ``-ML'' suffix for comparison models.}
\label{fig:jnr_curve}
\end{figure*}

\rev{Fig.~\ref{fig:jnr_curve} examines only the 8,887 training-listed test records, whereas Fig.~\ref{fig:composition_confusion} includes all 14,220 records. Their rankings are therefore not contradictory: the JNR curves emphasize familiar compositions, while the confusion matrices also expose errors on held-out combinations. The model settings remain fixed across these views.}

\rev{The low-JNR three-component comparison in Table~\ref{tab:listed_k3_lowjnr} identifies a clear cost of the proposed training recipe. At $-15$ dB, SPOC-Net attains 10.9\% accuracy, compared with 40.4\% and 43.3\% for MSFF-KAN-ML and TSFANet-ML. The gap narrows as JNR increases. This suggests that the advantage on new compositions does not remove the need for better weak-component recognition on familiar ones. Exposure to measured mixtures is one plausible contributor to the references' advantage in this region, but the present comparison does not separate that factor from architectural and inference differences.}

\begin{table}[t]
\centering
\caption{Three-Component Accuracy on Training-Listed Compositions at Low JNR}
\label{tab:listed_k3_lowjnr}
\setlength{\tabcolsep}{2.5pt}
\scriptsize
\begin{tabular}{crrr}
\toprule
JNR (dB) & SPOC-Net (\%) & MSFF-KAN-ML (\%) & TSFANet-ML (\%)\\
\midrule
$-20$ & 0.2 & 2.9 & 20.0\\
$-15$ & 10.9 & 40.4 & 43.3\\
$-10$ & 61.0 & 80.4 & 76.6\\
$-5$ & 92.1 & 96.4 & 93.9\\
\bottomrule
\end{tabular}
\end{table}

\subsection{Composition-Level and Primitive-Level Results}

\begin{figure*}[!t]
\centering
\includegraphics[width=0.329\textwidth]{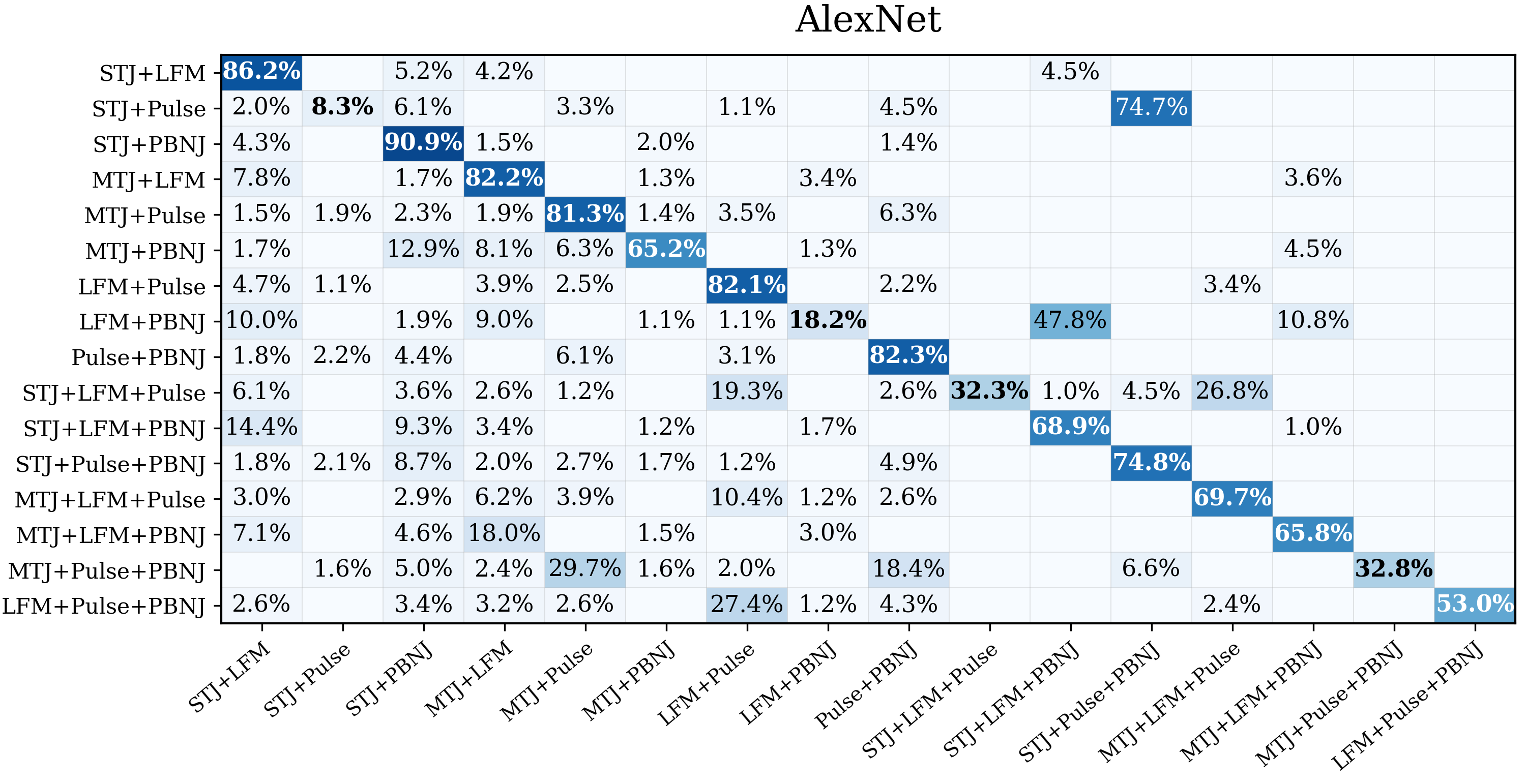}
\hfill
\includegraphics[width=0.329\textwidth]{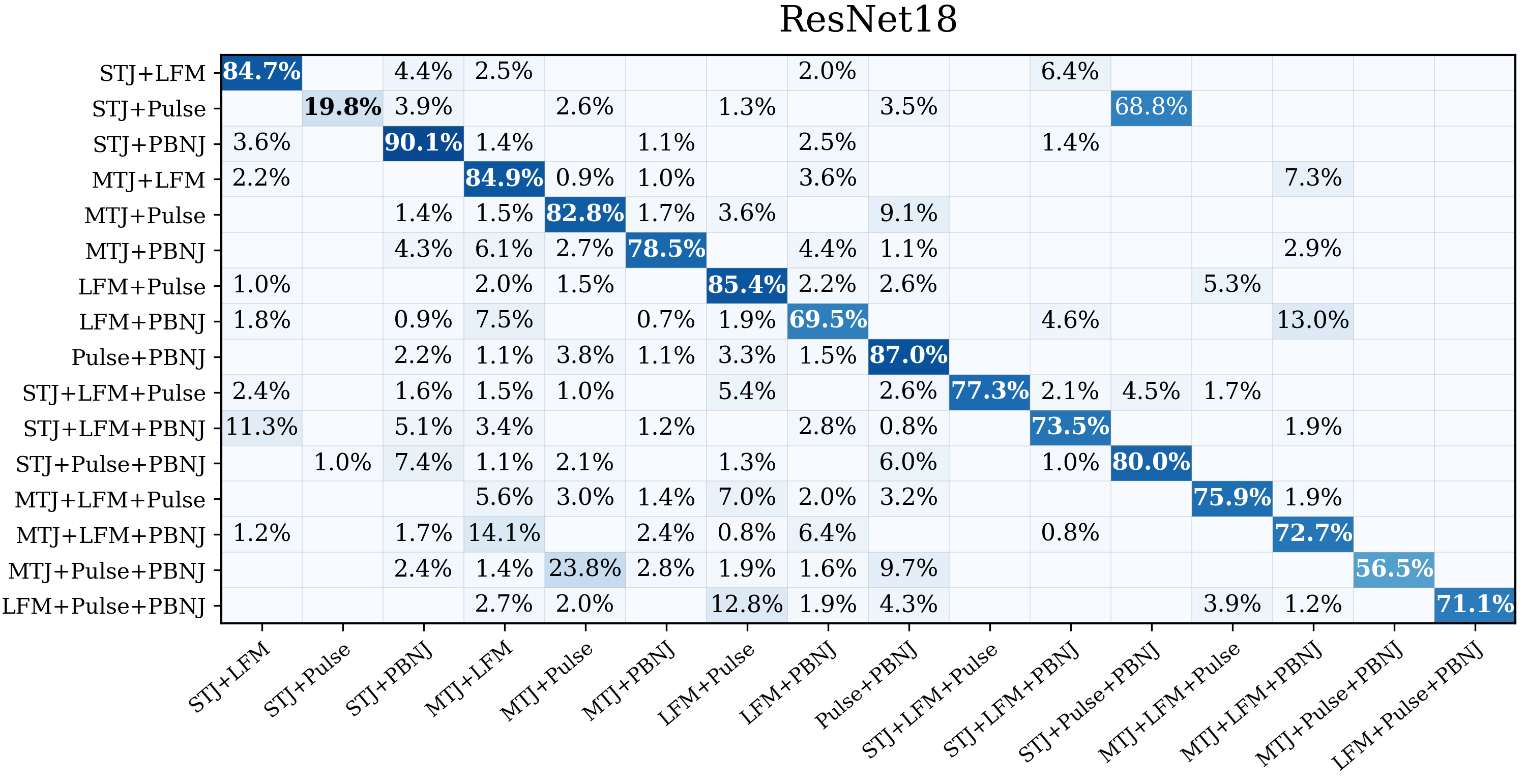}
\hfill
\includegraphics[width=0.329\textwidth]{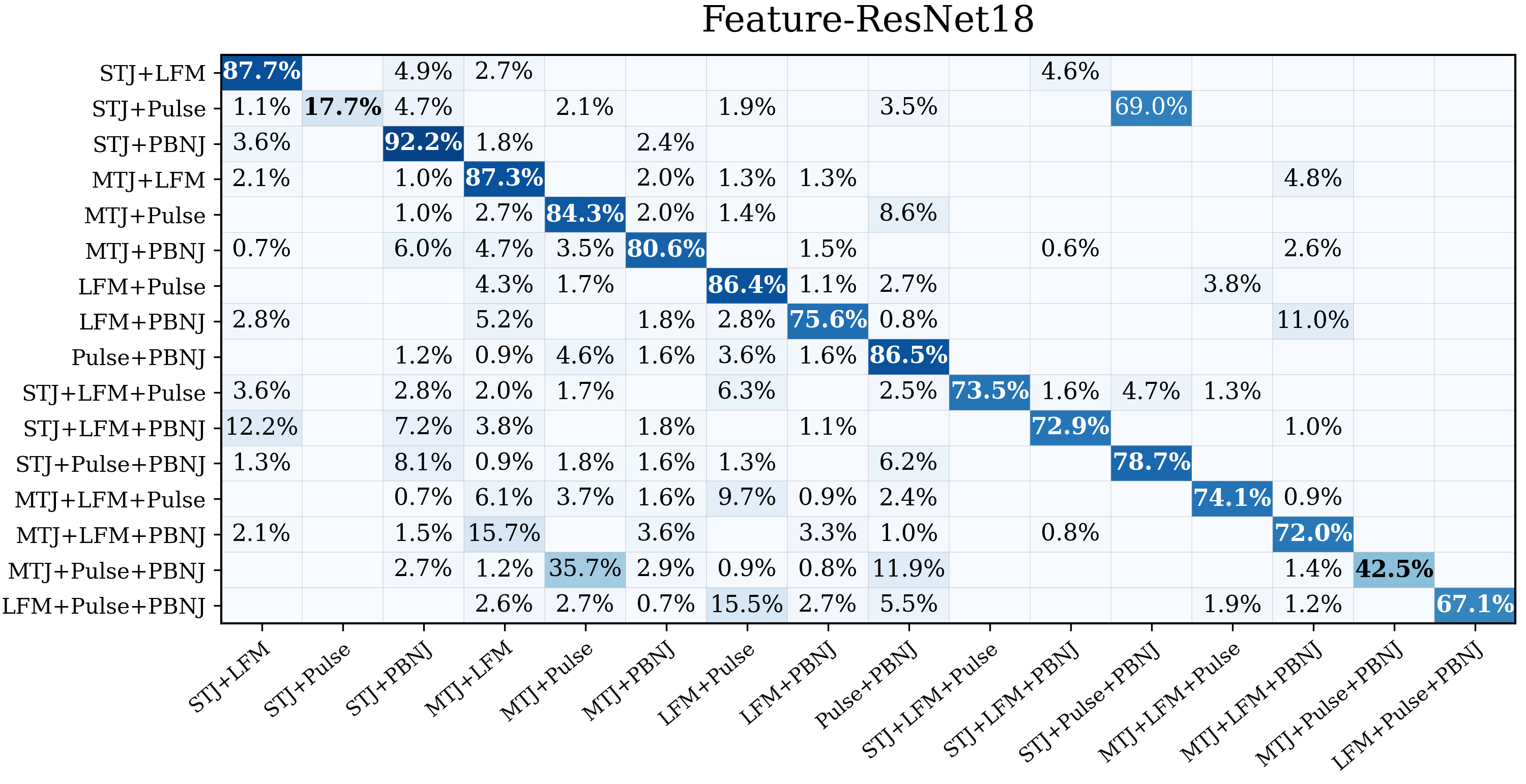}

\par\vspace{0.5ex}

\includegraphics[width=0.329\textwidth]{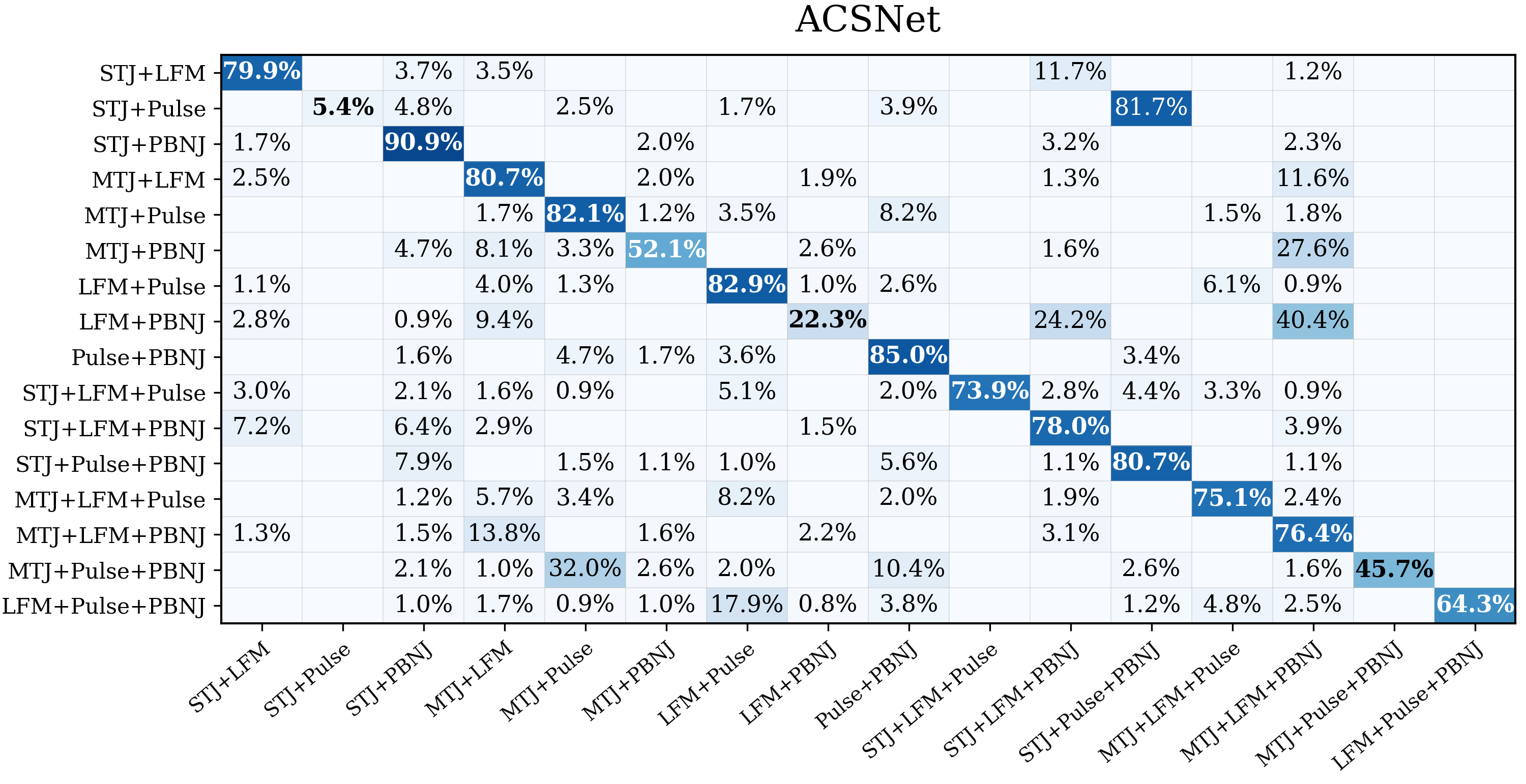}
\hfill
\includegraphics[width=0.329\textwidth]{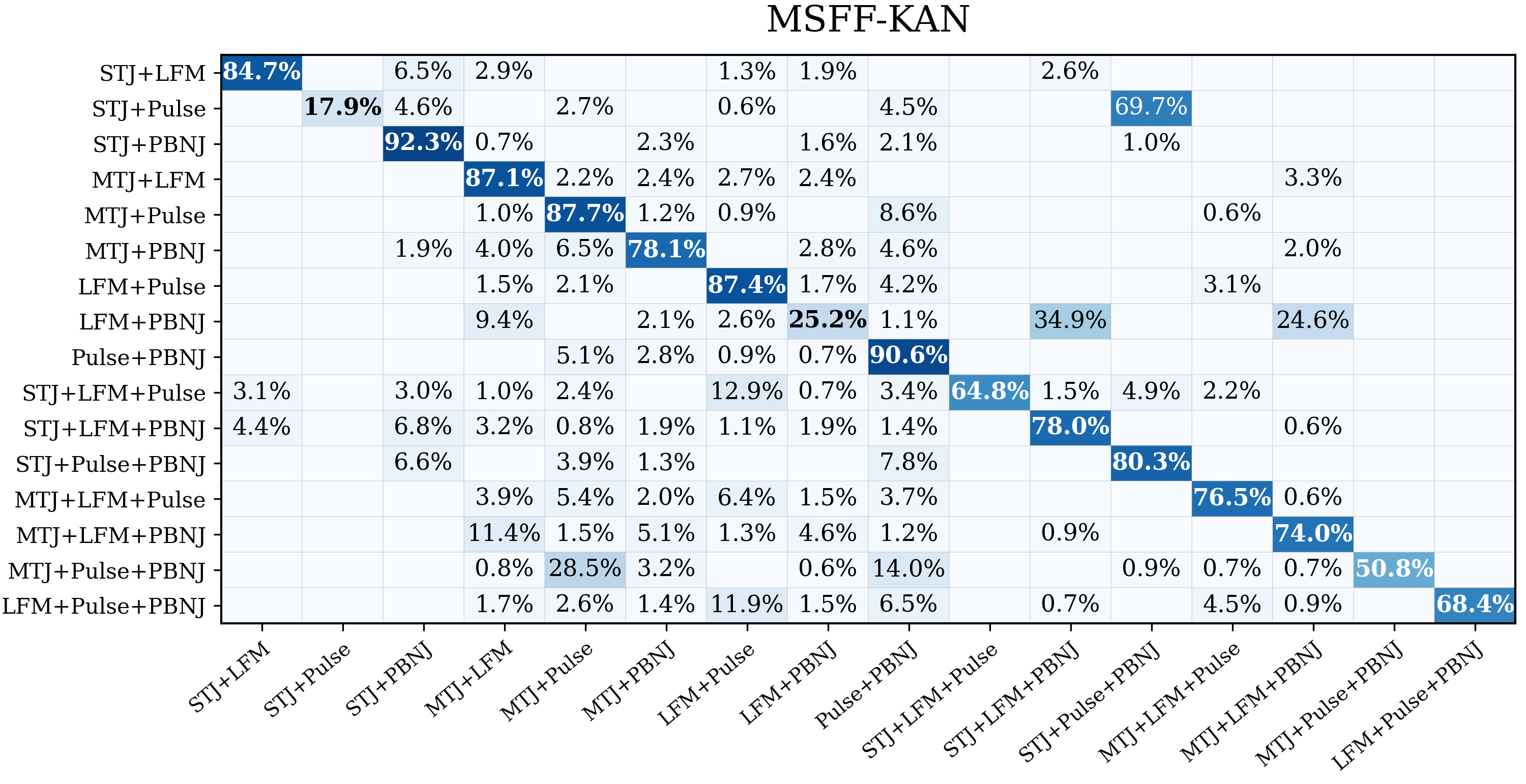}
\hfill
\includegraphics[width=0.329\textwidth]{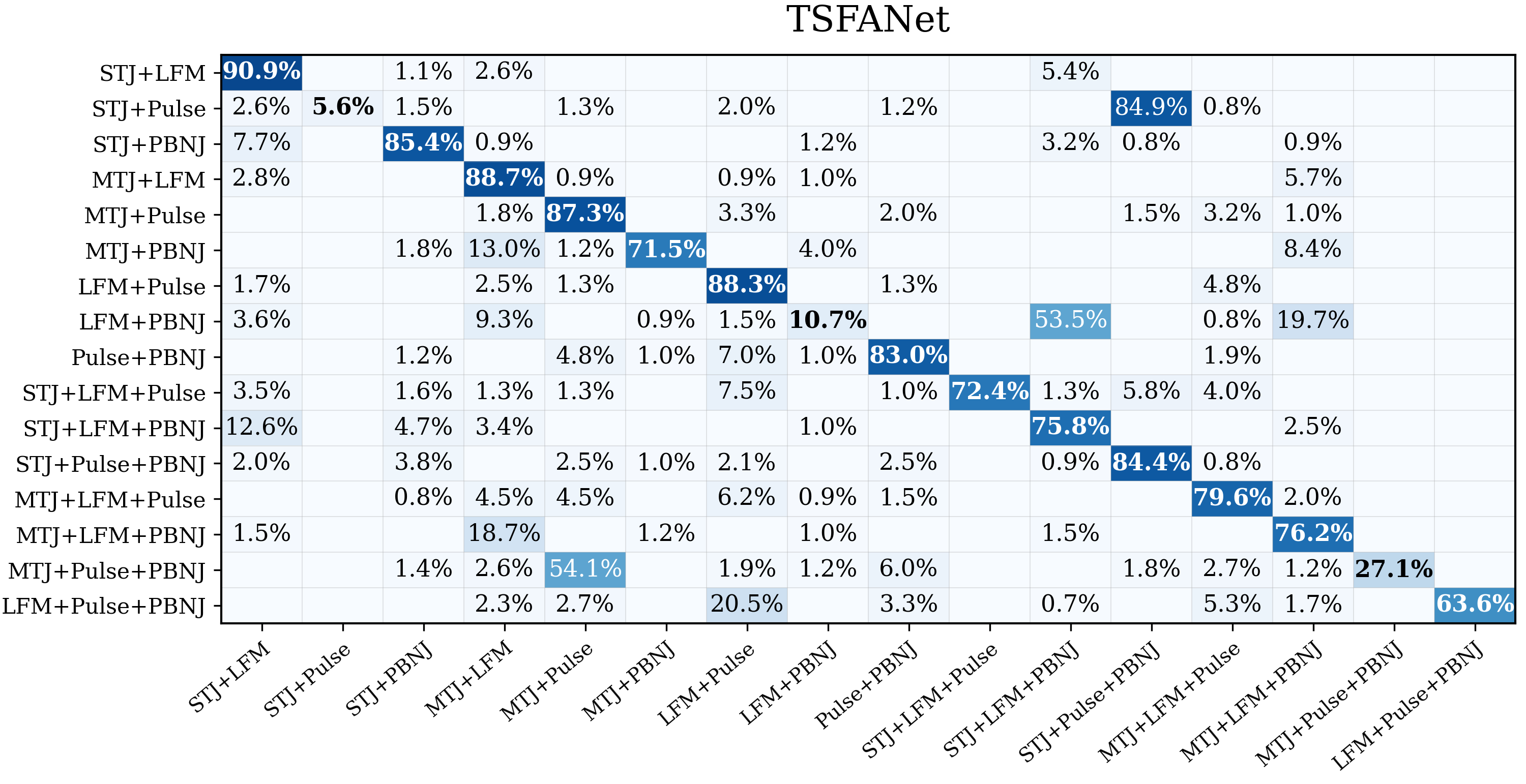}

\par\vspace{0.5ex}

\makebox[\textwidth][c]{%
\includegraphics[width=0.329\textwidth]{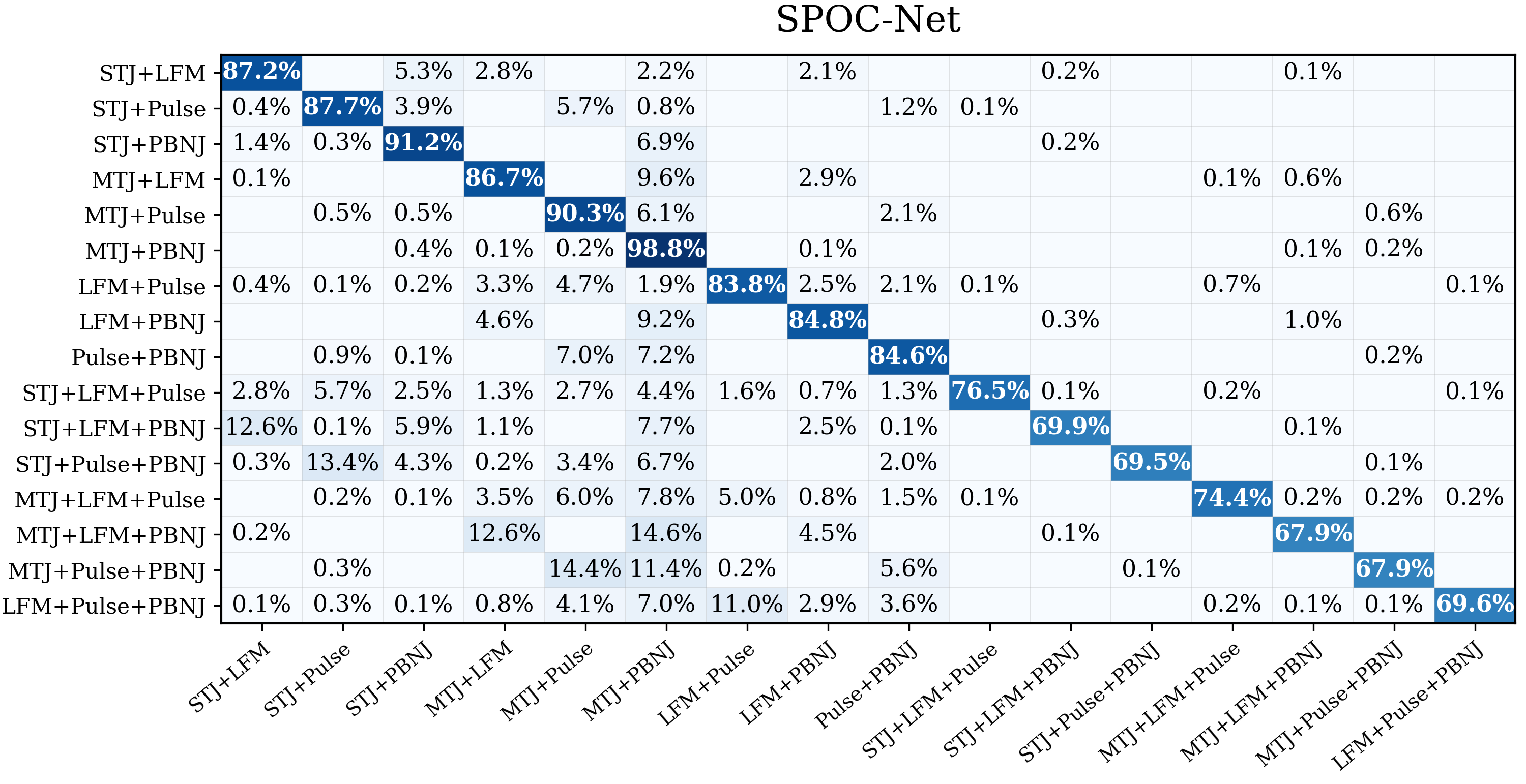}}

\caption{Row-normalized composition confusion matrices on the full measured mixed set. The upper row contains AlexNet-ML, ResNet18-ML, and Feature-ResNet18-ML. The middle row contains ACSNet-ML, MSFF-KAN-ML, and TSFANet-ML. The lower panel contains SPOC-Net. All 16 two-component and three-component classes and all tested JNR values are included. The artwork omits the ``-ML'' suffix.}
\label{fig:composition_confusion}
\end{figure*}

\rev{The row-normalized confusion matrices in Fig.~\ref{fig:composition_confusion} show whether the partition averages conceal composition-specific failures. The reference models exhibit pronounced errors on several held-out rows, whereas SPOC-Net retains a stronger diagonal there. The numerical discussion below and Table~\ref{tab:composition_results} complement the matrices by identifying the relevant compositions explicitly.}

\rev{Table~\ref{tab:composition_results} also shows that held-out status alone does not determine difficulty. For example, SPOC-Net recognizes MTJ+PBNJ with 98.76\% accuracy, while the three held-out three-component sets range from 67.91\% to 76.49\%. STJ+PTJ provides a particularly clear contrast with the references: SPOC-Net reaches 87.7\%, whereas the next-best result in Fig.~\ref{fig:composition_confusion} is 19.8\% for ResNet18-ML. This example demonstrates successful recombination of known labels, but the lower three-component accuracies prevent a claim of uniformly reliable transfer. The 94.13\% micro-F1 for MTJ+PTJ+PBNJ alongside its 67.91\% exact-set accuracy further illustrates how recovering most labels can conceal frequent incomplete sets.}

\begin{table}[t]
\centering
\caption{SPOC-Net Results on the Six Held-Out Compositions}
\label{tab:composition_results}
\begin{tabular}{lrrr}
\toprule
Composition & \rev{$N_{\mathrm{rec}}$} & Exact-set (\%) & Micro-F1 (\%)\\
\midrule
$\mathrm{STJ}+\mathrm{PTJ}$ & 889 & 87.74 & 93.51\\
$\mathrm{MTJ}+\mathrm{PBNJ}$ & 890 & 98.76 & 99.47\\
$\mathrm{LFMJ}+\mathrm{PBNJ}$ & 889 & 84.81 & 92.77\\
$\mathrm{STJ}+\mathrm{LFMJ}+\mathrm{PTJ}$ & 889 & 76.49 & 89.84\\
$\mathrm{MTJ}+\mathrm{PTJ}+\mathrm{PBNJ}$ & 888 & 67.91 & 94.13\\
$\mathrm{LFMJ}+\mathrm{PTJ}+\mathrm{PBNJ}$ & 888 & 69.59 & 90.24\\
\bottomrule
\end{tabular}
\end{table}

\rev{The primitive-level results in Table~\ref{tab:primitive_results} distinguish missed labels from spurious ones. STJ, LFMJ, and PTJ have precision above 98.9\%, but their lower recall indicates conservative predictions. LFMJ has the lowest recall, consistent with the difficulty of retaining a weak chirp in an overlapping mixture. MTJ exhibits the opposite imbalance: its recall exceeds its precision, indicating more false activations relative to missed detections. Confusion between narrowband patterns and structured regions of other signals is a possible explanation, but these aggregate primitive scores do not establish which competing type causes each error.}

\begin{table}[t]
\centering
\caption{Primitive-Wise Performance on the Full Mixed Set}
\label{tab:primitive_results}
\begin{tabular}{lrrr}
\toprule
Primitive & Precision (\%) & Recall (\%) & F1 (\%)\\
\midrule
STJ & 98.99 & 90.29 & 94.44\\
MTJ & 85.34 & 95.72 & 90.23\\
LFMJ & 99.59 & 87.62 & 93.22\\
PTJ & 99.86 & 90.50 & 94.95\\
PBNJ & 92.57 & 90.00 & 91.26\\
\bottomrule
\end{tabular}
\end{table}

\subsection{Cardinality Errors and Practical Limitations}

\rev{Table~\ref{tab:cardinality_confusion} identifies underestimation of component count as the dominant cardinality error: 28.86\% of true three-component records are assigned two components, whereas only 0.52\% of two-component records are assigned three. Moreover, three-component cardinality accuracy (71.14\%) is close to exact-set accuracy (70.83\%). On this subset, selecting the correct number of types therefore almost always coincides with selecting the correct identities. The principal remaining challenge is recognizing that another component is present, rather than frequently choosing the wrong three-type combination.}

\begin{table}[t]
\centering
\caption{Row-Normalized Cardinality Confusion}
\label{tab:cardinality_confusion}
\begin{tabular}{lrr}
\toprule
True cardinality & Predicted 2 (\%) & Predicted 3 (\%)\\
\midrule
2 & 99.48 & 0.52\\
3 & 28.86 & 71.14\\
\bottomrule
\end{tabular}
\end{table}

\rev{The most frequent set errors in Table~\ref{tab:top_errors} reinforce this interpretation. Every listed prediction is a two-component subset of a three-component target, with the remaining components identified correctly. In these cases, the missing label is PBNJ or LFMJ. Together with the low-JNR and recall results, these omissions are consistent with insufficient evidence for an additional type. The error counts describe the observed failures; they do not by themselves establish the received power of each omitted component.}

\begin{table}[t]
\centering
\caption{Most Frequent Set Errors}
\label{tab:top_errors}
\setlength{\tabcolsep}{3pt}
\begin{tabular}{p{0.40\columnwidth}p{0.35\columnwidth}r}
\toprule
True set & Predicted set & Count\\
\midrule
$\{\mathrm{MTJ},\mathrm{LFMJ},\mathrm{PBNJ}\}$ & $\{\mathrm{MTJ},\mathrm{PBNJ}\}$ & 130\\
$\{\mathrm{MTJ},\mathrm{PTJ},\mathrm{PBNJ}\}$ & $\{\mathrm{MTJ},\mathrm{PTJ}\}$ & 128\\
$\{\mathrm{STJ},\mathrm{PTJ},\mathrm{PBNJ}\}$ & $\{\mathrm{STJ},\mathrm{PTJ}\}$ & 119\\
$\{\mathrm{MTJ},\mathrm{LFMJ},\mathrm{PBNJ}\}$ & $\{\mathrm{MTJ},\mathrm{LFMJ}\}$ & 112\\
$\{\mathrm{STJ},\mathrm{LFMJ},\mathrm{PBNJ}\}$ & $\{\mathrm{STJ},\mathrm{LFMJ}\}$ & 112\\
\bottomrule
\end{tabular}
\end{table}

\rev{Taken together, the results support singleton-based IQ composition as a useful route to recognizing untrained combinations under the reported acquisition conditions. The method retains competitive accuracy on training-listed mixtures and transfers more effectively than the reference procedures on the chosen held-out split. The evidence supports this system-level conclusion, not a separate causal claim for each network module.}

\rev{The main limitation is low-JNR recognition of a third component. Fixed-range STFT clipping and subsequent feature compression can suppress weak patterns; the high-resolution branch and paired objective are designed to address this issue, but their individual effects are not isolated by the reported results. The conducted platform provides repeatable RF superposition but does not reproduce outdoor source-dependent propagation, antenna-pattern changes, or multipath. Relative to the linear composer, the physical path can introduce transmitter and receiver responses, combiner effects, gain variation, and front-end impairments. Their individual contributions are not quantified here, so the results establish performance on this hardware setup rather than general robustness to every RF impairment.}

\rev{The present evaluation also uses one fixed composition partition and does not report uncertainty across repeated training runs. Generalization to other partitions therefore remains to be assessed. The decoder is limited to the five defined primitives and valid two-component and three-component sets; it is not a complete interference monitor with no-interference, singleton, and unknown-type decisions. Extending those decisions, evaluating additional composition splits, and quantifying run-to-run variability are directions for future work.}

\section{Conclusion}
\label{sec:conclusion}

\rev{This paper presented SPOC-Net for identifying the component types of mixed GNSS jamming when measured single-component records are the only physical observations used for gradient optimization. Its online IQ composer generates labeled mixtures during training, while component queries, a high-resolution cardinality branch, and structured decoding combine identity and set-size evidence at inference. Measured mixtures of training-listed compositions remain available for model selection and calibration, and six other compositions are excluded from every development decision. On 14,220 independently acquired conducted mixed records, the method achieves 80.69\% exact-set accuracy and 92.84\% micro-F1. Its held-out accuracy of 80.89\% exceeds the strongest reference by 18.77 percentage points under the reported training and inference protocols. The results support recognition of new combinations of known jamming types without measured-mixture gradient training. Weak-component omissions at low JNR, the fixed composition split, and the conducted acquisition setting delimit that conclusion.}

\end{document}